\documentclass{article}

\usepackage[preprint]{neurips_2026}

\usepackage[utf8]{inputenc} 
\usepackage[T1]{fontenc}    
\usepackage{hyperref}       
\usepackage{url}            
\usepackage{booktabs}       
\usepackage{amsfonts}       
\usepackage{nicefrac}       
\usepackage{microtype}      
\usepackage{xcolor}         

\usepackage{graphicx}
\usepackage{subcaption}
\usepackage{amsmath}
\usepackage{amssymb}
\usepackage{mathtools}
\usepackage{amsthm}
\usepackage{cleveref}
\usepackage[textsize=tiny]{todonotes}
\usepackage{caption}
\usepackage{minted}
\usepackage{inconsolata}
\usepackage[most]{tcolorbox}
\usepackage{multirow}
\usepackage{placeins}
\usepackage{wrapfig}
\usepackage{arydshln}
\definecolor{codebg}{HTML}{FBFBFD}
\definecolor{codeframe}{HTML}{D0D7DE}

\definecolor{citeblue}{HTML}{4962A5}
\hypersetup{
    colorlinks=true,
    citecolor=citeblue,
    linkcolor=citeblue,
    urlcolor=citeblue,
}

\usepackage{titletoc}

\newcommand{\AppendixToC}{%
  \startcontents[appendix]%
  \printcontents[appendix]{}{1}{%
    \section*{Appendix}%
    \vspace{-0.75em}%
    \hrule%
    \vspace{0.5em}%
  }%
  \vspace{1em}
  \hrule
  \vspace{1em}
}

\tcbset{
  codebox/.style={
    colback=codebg,
    colframe=codeframe,
    boxrule=0.4pt,
    arc=2mm,
    left=2mm,right=2mm,top=1mm,bottom=1mm,
    enhanced, breakable,
  }
}

\theoremstyle{plain}
\newtheorem{theorem}{Theorem}[section]

\theoremstyle{definition}
\newtheorem{definition}[theorem]{Definition}

\theoremstyle{remark}

\title{Generating Financial Time Series by Matching Random Convolutional Features}

\author{%
  Konrad J. Mueller\thanks{Equal contribution} \,\thanks{Corresponding author: \texttt{k.mueller23@imperial.ac.uk}} \\
  Imperial College London\\
  JPMorgan Chase \& Co.\thanks{Opinions expressed in this paper are those of the authors, and do not necessarily reflect the view of JPMorgan Chase~\&~Co.}
  \And
  Nikita Zozoulenko$^*$ \\
  Imperial College London
  \AND
  Ben Wood \\
  JPMorgan Chase \& Co.\textsuperscript{‡}
  \And 
  Thomas Cass \\
  Imperial College London
  \And
  Lukas Gonon \\
  Imperial College London \\
  University of St. Gallen
}

\begin{document}

\maketitle

\begin{abstract}
Generating realistic financial time series is challenging as training data
is often limited to a single historical path.
With such scarce data, overfitting is hard to avoid,
especially under adversarial training where
a trained discriminator can memorize the training samples.
To mitigate this,
recent approaches train generators
to minimize the discrepancy between
untrained feature representations of real and generated time series.
In these works, the feature maps are based on path signatures,
which can fail to capture relevant time series properties
at tractable truncation depths.
In this work, we instead train generators by matching random convolutional
features of real and generated time series.
Existing random convolutional feature maps, such as Rocket and Hydra,
have been shown to provide informative representations of real-world time series,
but cannot supervise generative models because they are non-differentiable.
We introduce
\textbf{\texttt{SOCK}}
(\textit{\textbf{SO}ft \textbf{C}ompeting \textbf{K}ernels}),
a fully differentiable random convolutional feature map,
suited to train generative time series models.
We show that generators trained by matching random \texttt{SOCK} features
outperform signature and diffusion
baselines on most evaluation metrics
across a wide range of small-sample financial datasets.
We further demonstrate \texttt{SOCK}'s expressiveness
on two-sample hypothesis testing and
time series classification tasks,
where \texttt{SOCK} matches or outperforms existing
unsupervised feature maps.
\end{abstract}

\vspace{-1em}
\begin{center}
    \small Code: \url{https://github.com/konmue/sock}
\end{center}

\section{Introduction}\label{sec:intro}
Training generative time series models with limited real-world data
is a fundamental challenge across many domains.
In finance, time series datasets are often small in sample size:
for instance, eight years of daily trading data provides about 2,000 data points.
Because historical data is limited,
synthetic time series are important for many applications,
including estimation of risk measures \citep{cont_2025_tailg}
and policy optimization for investment
\citep{mariani_2019_pagan, koshiyama_2021_gener}
or hedging problems
\citep{wiese_2019_deepa, buehler_2020_a, cont_2025_datad}.
These applications require sampling future paths
conditional on recent observations.
We therefore study time series generation with two practical constraints:
(i) only a single historical path is available for training, and
(ii) generation is conditional on the $q$ most recent observations.

Training generative models adversarially is often unstable with such limited data
and can drive the generator to memorize training samples \citep{karras_2020_train, issa_2023_nonad}.
Instead, recent works propose training time series generators
via feature matching under a fixed feature map
\citep{ni_2022_sigwa, 2024NiklasGononUniversalrandomisedsignaturesgenerative}.
They fit a generative model $p_\theta$ to minimize
the mean feature discrepancy
\begin{equation}\label{eq:feature_matching_loss}
    \lVert\mathbb{E}_{X \sim p}[f(X)]
    - 
    \mathbb{E}_{X \sim p_\theta}[f(X)] \rVert_2^2 ,
\end{equation}
where $X = (X_1, \dots, X_T)$
is the $\mathbb{R}^d$-valued time series of length $T$,
and $p$ denotes the training data distribution.
In contrast to (adversarial) feature-matching
\citep{salimans_2016_impro,mroueh_2017_mcgan},
the feature map ${f: \mathbb{R}^{T \times d} \to \mathbb{R}^n}$
is fixed rather than learned,
and thus cannot overfit to the training data.
This makes such non-adversarial feature-matching
well suited to the small-sample regime.

Motivated by rough path theory \citep{1998RoughPathTheory},
prior work chooses the path statistics~$f(X)$
from the family of signature-based path-to-vector transformations.
While the signature transform has strong theoretical properties,
other, more heuristically motivated, unsupervised feature maps
empirically outperform truncated signature features
on small-sample time series classification problems \citep{2024BakeOff}.
On these tasks,
random convolutional feature maps such as
Rocket \citep{2020Rocket,2022MultiRocket}
and Hydra \citep{2023Hydra}
often yield the strongest classifiers.

In this paper, we show that the strong classification performance
of random convolutional features transfers to generative modeling.
Their randomness is especially useful for feature matching:
the feature map can be resampled during generator training,
so supervision does not rely on a single fixed representation.
Existing random convolutional feature maps are not differentiable,
so they cannot directly supervise generator training.
We address this by introducing a new differentiable random convolutional feature map.
Our contributions are:
\begin{enumerate}
    \item \textbf{Random convolutional feature matching for conditional generation.}
    We train conditional time series generators
    by matching random convolutional feature statistics
    of real and generated paths.
    During training, we resample the random feature map,
    so the generator is supervised by a distribution of feature maps.

    \item \textbf{Differentiable random convolutional features.}
    We introduce \texttt{SOCK},
    a random convolutional feature map for multivariate time series
    that is fully differentiable w.r.t.\ the input path.
    \texttt{SOCK} builds on Hydra's random kernel competitions,
    but differs in how it preprocesses and convolves inputs,
    and pools the convolutional responses in a differentiable way.
    
    \item \textbf{Evaluation under limited-data training.}
    We evaluate generator training techniques in a practical setting:
    training on a single historical path and evaluating out-of-sample.
    Across synthetic and real financial datasets,
    generators trained by matching random \texttt{SOCK} features
    achieve stronger overall performance
    than signature baselines and Diffusion-TS \citep{yuan_2024_diffu}.
    We further validate \texttt{SOCK}'s expressiveness
    on two-sample hypothesis testing and time-series classification tasks.
\end{enumerate}
 
\section{Background and related work}

\paragraph{Signature-based path features}
The signature transform
maps a continuous-time path
to an infinite collection of iterated integrals,
which can be viewed as statistics of the path.
In practice, we can truncate the signature to compute
a finite feature representation of
a discrete-time path $X \in \mathbb{R}^{T \times d}$,
which involves augmenting and interpolating $X$ \citep{2025SignatureMethodsInMachineLearning, 
2020SignatureFeaturesWithTheVisibilityTransformation,2019KernelsForSequentiallyOrderedData}.
Due to the polynomial growth of the signature in terms of the state-space dimension,
the randomized signature is often preferred
\citep{2021ExpressivePowerOfRandomizedSignatures,2023NeuralSignatureKernels,2025InfiniteDimensionalMahalanobisDistance}.

\paragraph{Non-adversarial feature matching for time series}
Matching expected feature statistics 
is a common objective for training generative models
across various data modalities.
Existing methods either learn the feature map $f$ (or a kernel)
jointly with the generator
\citep{salimans_2016_impro, mroueh_2017_mcgan,li_2017_mmd, sutherland_2017_gener},
or fix $f$ prior to training
\citep{li_2015_gener,dziugaite_2015_train}.
\citet{ni_2022_sigwa} apply such non-adversarial feature matching
to time series generation by using the truncated signature
transform for $f$.

\citet{ni_2022_sigwa} study unconditional generation,
whereas many financial applications require generating
future paths conditional on the recent past.
\citet{liao_2024_sigwa} train such conditional generators
with signature feature matching.
Instead of directly applying \cref{eq:feature_matching_loss}
to conditional samples,
\citet{liao_2024_sigwa} match conditional feature means.
They estimate the conditional feature mean under the generated distribution
by sampling multiple futures for each past segment,
and approximate the real conditional mean 
with a linear predictor from the past segment's features
(details in \cref{app:subsec:baseline_details}).
\citet{2024NiklasGononUniversalrandomisedsignaturesgenerative}
extend both the unconditional and conditional feature-matching objectives
to randomized signature features.
Empirically, this line of work
reports strong performance of signature-based feature matching
on small-sample financial datasets.
For instance, \citet{liao_2024_sigwa} report improvements
over adversarial baselines
\citep{esteban_2017_realv, yoon_2019_times}
across several distributional metrics and financial datasets.
These results motivate our work on training generators
with untrained and randomized feature maps.

\paragraph{Random convolutional time series features}
In the small-sample time series classification literature \citep{2024BakeOff}, the best performing classifiers most often take the form of a simple linear classifier on top of an untrained (unsupervised) high-dimensional feature map.
For this purpose, random convolutional feature maps,
such as the Rocket and Hydra families
\citep{2020Rocket, 2021MiniRocket, 2022MultiRocket, 2023Hydra, 2024HighlyScalableTimeSeriesClassificationForVeryLargeDatasets},
perform particularly well.
These 
compute features by convolving
the input path with many random kernels
and pooling the convolutional responses over time
via non-differentiable operations such
as $\mathrm{argmax}$ or indicator functions.

\section{Random convolutional feature matching}\label{sec:conditional_generation}

\subsection{Conditional time series generation from limited data}\label{subsec:problem_setting}

In financial applications of time series generative models,
one often observes only a single realized path of the time series,
but needs to simulate many plausible future scenarios conditional on the
recent past.
We study time series generation in this practically important setting.
Concretely, our goal is to train a conditional generator that,
given the most recent $q$ observations of a time series,
produces a distribution over the next $T$ observations.
Writing $x^-\in\mathbb{R}^{q\times d}$ for these recent observations,
the generator defines a conditional distribution $p_\theta(\cdot\mid x^-)$
over length-$T$ futures in $\mathbb{R}^{T\times d}$.

The practical constraints determine how we train and evaluate the generator.
During training, only a single historical path
$X_{1:H}\in\mathbb{R}^{H\times d}$ of length $H$ is available.
We evaluate the generator on the length-$N$ out-of-sample continuation of this path,
denoted $X_{H+1:H+N}$.
At each $t \geq H$, we condition on the recent history $x_t^-$
to generate a future $\hat x_t^+$ and compare it to the realized future $x_t^+$:
\begin{equation*}
x_t^- := X_{t-q+1:t},
\qquad
x_t^+ := X_{t+1:t+T},
\qquad
\hat x_t^+ \sim p_\theta(\cdot\mid x_t^-).
\end{equation*}
We then form the joined past--future segments
$x_t^-\oplus\hat x_t^+$ and $x_t^-\oplus x_t^+$ for each evaluation time,
where $\oplus$ denotes concatenation along the time axis.
We aggregate these segments over evaluation times $t$
into empirical distributions of generated and realized segments.
The generator is then evaluated by comparing these empirical path distributions;
see \cref{tab:metrics} for evaluation metrics.

On synthetic datasets, where the data-generating law is known,
we sample $J$ independent continuations $X^{(j)}_{H+1:H+N}$ for $j=1,\dots,J$.
For each continuation, we form generated and realized joined segments as above
and then aggregate those over both continuations $j$ and times $t$.
Multiple continuations increase the sample size of the
empirical path distributions and reduce evaluation noise.

\subsection{Training conditional time series generators with random feature matching}\label{subsec:nonadversarial_fm}

Training the conditional generator $p_\theta$ is challenging
due to the limited training data.
Adversarial methods are particularly prone to overfitting in this setting,
since the training data contains only one realized future $x_t^+$
for each conditioning segment $x_t^-$.
Instead, we train the generator $p_\theta$ by matching
random convolutional features of real and generated paths.
Our method builds on the signature matching approach of \citet{ni_2022_sigwa},
but differs in three respects.
First, we replace signature features with random convolutional features.
Second, we train against repeated draws of the random feature map,
rather than a single fixed feature map.
Third, we train a conditional generator
by applying the feature map to joined past--future segments
$x_t^- \oplus x_t^+$ and $x_t^- \oplus \hat{x}_t^+$.

Let $f_\psi:\mathbb{R}^{(q+T)\times d}\to\mathbb{R}^{d_f}$ be a feature map
with random parameters $\psi\sim\Pi$.
We define our choice of $f_\psi$, the \texttt{SOCK} feature map,
in \cref{sec:sock}.
We train the generator parameters $\theta$ with the objective
\begin{equation}
\begin{aligned}
\min_\theta \quad
\mathbb{E}_{\psi\sim\Pi}\Bigl[
\Bigl\|
&\mathbb{E}_{t}
\bigl[
    f_\psi(x_t^- \,\oplus\, x_t^+)
\bigr]
-
\mathbb{E}_{\substack{
    t,\;\hat{x}_t^+\sim p_\theta(\cdot\mid x_t^-)
}}
\bigl[
    f_\psi(x_t^- \,\oplus\, \hat{x}_t^+)
\bigr]
\Bigr\|_2^2
\Bigr] ,
\end{aligned}
\label{eq:random_feature_matching_objective}
\end{equation}
where $t$ is sampled uniformly from the valid training indices
$\{q,\dots,H-T\}$.
The outer expectation over $\psi$ trains the generator to match feature means
across a distribution of feature maps.\footnote{
Similar objectives using randomly initialized neural networks
have been used to directly optimize synthetic image data
\citep{zhao_2023_dataset}.}
This reduces the risk of supervisory blind spots of any single feature-map draw:
discrepancies that are invisible to one draw of $f_\psi$
may be detected by other independent draws.

In practice, we estimate the inner expectations by sampling
$B$ time indices $t_1,\dots,t_B$ and generating a single
continuation $\hat x_{t_i}^+\sim p_\theta(\cdot\mid x_{t_i}^-)$
for each context $x_{t_i}^-$.
We estimate the outer expectation over $\psi$ with one
feature-map draw at a time and keep this draw fixed for $R$ optimization steps
before resampling. The main additional cost of resampling is recomputing
features of the real path segments, which could otherwise be cached for a fixed
feature map.
Resampling only every $R$ steps amortizes this cost while still
exposing the generator to many independent feature maps during training,
which can substantially improve the performance of the trained generator
(Appendix~\ref{app:subsec:ablations}).

The success of this approach depends on the random feature map $f_\psi$.
Its feature means must reveal relevant discrepancies between real and generated
paths,
and the feature map must be differentiable with respect to its input so that
gradients pass through generated paths to $\theta$.
Existing random convolutional features are highly expressive, but their pooling
operations are typically non-differentiable.
We therefore introduce a differentiable random convolutional feature map next.

\subsection{Differentiable random convolutional features}\label{sec:sock}

We introduce \texttt{SOCK}, a differentiable random convolutional feature map
for multivariate time series.
\texttt{SOCK} maps a path to features in three steps:
(i) preprocessing and augmentations,
(ii) grouped random convolutions,
and (iii) differentiable temporal pooling.
The feature map is based on Hydra~\citep{2023Hydra},
which first showed that competitions between random convolutional kernels
provide useful features for classifying real-world time series.
\texttt{SOCK} builds on the idea of kernel competitions,
but differs from Hydra in several respects.
Most importantly, \texttt{SOCK} supports multivariate input series
and uses a differentiable temporal pooling.

\textbf{Preprocessing.}
We augment, normalize, and randomly project the input path
$X \in \mathbb{R}^{T \times d}$ to obtain
\begin{align*}
    Y_t = P \; \mathcal{N}\!\left(\mathcal{A}(X)\right)_t .
\end{align*}
Here $\mathcal{A}: \mathbb{R}^{T \times d} \to \mathbb{R}^{T \times d'}$
is an augmentation map,
$\mathcal{N}$ denotes per-channel centering and scaling using statistics
fitted on the augmented training paths,
and $P \in \mathbb{R}^{M \times d'}$ is a random Gaussian matrix
with $\ell_2$-normalized rows.
The augmentation map $\mathcal{A}$ preserves the original input path and appends
derived channels.
We make different augmentation choices in our generative and discriminative experiments.
We find that appending either the integrated ($\mathtt{int}$)
or differenced ($\mathtt{diff}$) path,
as well as elementwise positive and negative parts of each channel
($\mathtt{posneg}$) can sometimes improve performance.
See Appendices~\ref{app:subsec:sock_augmentations}
and \ref{app:subsec:ablations} for details and ablations.

\textbf{Grouped random convolutions.}
We partition the $M$ channels of $Y$ into $G$ groups of width $W$ and write
$Y^{(g)}\in\mathbb{R}^{T\times W}$ for group $g$.
Then, we convolve each $Y^{(g)}$ with $K$ random kernels
$\{w^{(g, k)}\}_{k = 1}^K$ of shape $W \times L$
at dilation $\delta \geq 1$.
For odd $L$, the convolution response at time $t$ is
\begin{equation*}
Z_t^{(g,k)} \; = \;
\sum\nolimits_{r=-(L - 1) /2}^{(L - 1)/2} \;
\big\langle w_{r}^{(g,k)},\, Y_{t+\delta r}^{(g)} \big\rangle ,
\end{equation*}
with zero padding when $t+\delta r \notin \{1,\dots,T\}$.
The kernels are sampled with i.i.d.\ Gaussian entries, then centered and
$\ell_1$-normalized, with full details given in Appendix~\ref{sec:code}.

This step largely follows Hydra,
including the default choices $L=9$ and $K=8$,
but differs in the group width.
Hydra uses channel-wise kernels ($W=1$),
whereas \texttt{SOCK} uses $W=2$, so that each $Z^{(g, k)}$
can combine two projected channels.
One difference between these choices
is how constant shifts in input channels affect $Z$.
Because each kernel's weights sum to zero,
$Y^{(g)}$ and $Y^{(g)} + c\mathbf{1}$ give the same response $Z^{(g,k)}$
away from zero-padding boundaries.
For $W=1$, this implies that each $Z^{(g, k)}$
is invariant to constant shifts in any channel of the
sequence $\mathcal{N}(\mathcal{A}(X))$.
For $W=2$, such per-channel shifts generally change $Z^{(g,k)}$,
because kernel centering only cancels joint shifts of the two channels in a group.

\textbf{Soft competing kernels.}
To map the response series $Z^{(g, \cdot)}$
into a feature vector, we follow Hydra's interpretation of the
$K$ parallel convolutions as a competition between kernels:
if $Z_t^{(g,k)} > Z_t^{(g,k')}$,
kernel $k$ outcompetes kernel $k'$ as it better matches the local path around time $t$.
Hydra summarizes these competitions
by counting how often each kernel `won'
and by accumulating the corresponding winning values~$Z_t^{(g,k)}$.
These features capture which random kernel directions best match the path over
time, but they depend on a hard selection over kernels and are thus not
differentiable in $Z$.\footnote{
Hydra uses
$\mathrm{argmax}$ for counts,
$\mathrm{argmin}$ for value features.
}
In \texttt{SOCK}, we represent the competition at each time step
by win probabilities
\begin{equation*}
P_t^{(g,k)}
=
\tfrac{\exp(Z_t^{(g,k)}/\tau)}
{\sum_{j=1}^K \exp(Z_t^{(g,j)}/\tau)} ,
\end{equation*}
which are differentiable with respect to the responses $Z_t^{(g, k)}$
for any temperature $\tau > 0$.
We find that several ways of aggregating the series $P^{(g,k)}$ over time
produce useful features.
In our experiments,
we use the \emph{soft-deviation} feature,
the standard deviation of the win probabilities over time:
\begin{equation*}
F^{(g,k)}
=
\mathtt{std}_{t=1,\dots,T}\bigl(P_t^{(g,k)}\bigr).
\end{equation*}
This statistic $F^{(g,k)}$ measures how much the
win probability of kernel $k$ in group $g$ varies over time.
We also consider differentiable generalizations of Hydra's count and value features,
obtained by averaging $P_t^{(g,k)}$ or $P_t^{(g,k)}Z_t^{(g,k)}$ over time
(Appendix~\ref{app:subsec:sock_pooling}).
Among individual pooling functions, soft-deviation performs best overall
(see right panel of \cref{fig:classification}).
Concatenating multiple statistics can further improve
performance in some settings (see discussion in Appendix~\ref{app:subsec:tsc}).

Following Hydra, we capture time series properties at multiple time scales
by applying the pipeline $Y\mapsto Z\mapsto F$
to the same $Y$ for different dilations
$\delta\in\mathcal D := \{2^e:\ e=0,1,\dots,e_{\max}\}$,
with independently sampled kernels per $\delta$.\footnote{The maximum exponent $e_{\max}$ depends on the input length (Appendix~\ref{sec:code}).}
We concatenate features across dilations
into a feature vector of dimension
${d_f = \lvert \mathcal{D} \rvert G K}$.
We provide additional details and pseudocode for \texttt{SOCK}
in Appendix~\ref{sec:code}.

\section{Conditional time series generation experiments}\label{sec:ts_gen}

We now consider the small-data,
conditional generation task of \cref{subsec:problem_setting}.
We compare
training generative models by matching random \texttt{SOCK} features
to signature-matching and diffusion baselines.

\paragraph{Synthetic datasets}

\begin{wraptable}{r}{0.55\textwidth} 
  \vspace{-1em}
  \centering
  \caption{Synthetic datasets and their key properties.}
  \label{tab:synth_datasets}
  \small
  \begin{tabular}{@{}lcll@{}}
    \toprule
    Dataset & $d$ & Order & Key phenomena \\
    \midrule
    \texttt{V1}  & 3 & 1        & strong autocorrelation ($\phi = 0.99$)\\
    \texttt{V10} & 3 & 10       & oscillatory autocorrelation \\
    \texttt{TGH} & 3 & 10       & skewed + heavy tail marginals \\
    \texttt{SV}  & 2 & 1        & cross-correlation (spot-vol), skew \\
    \texttt{FGN} & 3 & $\infty$ & long memory; $\mathcal{H} \in \{.05, .15, .25\}$ \\
    \bottomrule
  \end{tabular}
  \vspace{-0.75em}
\end{wraptable}
We first evaluate the generative modeling techniques on five synthetic datasets
spanning diverse time series characteristics (\cref{tab:synth_datasets}).
The datasets \texttt{V1} and \texttt{V10}
are sampled from vector autoregressive processes,
while \texttt{TGH} applies a Tukey $g$--$h$ marginal transform to \texttt{V10}
to induce skewness and heavy tails \citep{yan_2019_nonga}.
The \texttt{SV} dataset consists of the log-return process and log-variance process,
simulated under the stochastic volatility model of \citet{heston_1993_a}.
The dataset \texttt{FGN} is a multivariate fractional Gaussian noise process,
with independent components and component-specific Hurst exponents.
For each dataset, we sample a single training path of length $H = 2048$
(roughly eight years of trading days),
and $J = 2048$ independent out-of-sample continuations
of length $N = 2048$.
We repeat this protocol for five seeds, resampling all paths each time.
This reduces evaluation noise caused by the limited training data.

\paragraph{Real datasets}
We further evaluate on seven real financial datasets spanning multiple asset
classes.
We include four sector-based stock datasets
(\texttt{FI}, \texttt{PH}, \texttt{IT}, \texttt{ST}),
each a $4$-dimensional time series of daily log-returns
of four large-cap stocks from the same sector.
We further include two $3$-dimensional daily datasets:
\texttt{IDX}, comprising S\&P~500, VIX, and gold futures returns,
and \texttt{FX}, comprising EUR/USD, JPY/USD, and GBP/USD returns.
Finally, we include \texttt{CRY},
which consists of 5-minute returns of BTC and ETH.
The daily time series cover 2009--2025.
We use the first $8$ years for training
and the last $8$ years for evaluation
(details on datasets in Appendix~\ref{app:subsec:data}).

\paragraph{Rollout length}
We generate paths of length $T=64$ (roughly 3 months of trading days),
which is a relevant horizon for mid-term risk management applications.
This is more challenging than the short horizons
$T\in\{3, 5 ,10\}$ used in prior work
\citep{liao_2024_sigwa, 2024PathDevelopmentNetworkWithFiniteDimensionalLieGroup, 2024NiklasGononUniversalrandomisedsignaturesgenerative},
as even small modeling errors visibly compound over longer rollouts.
We use a short context length $q=5$ (one week of trading days),
to limit overfitting.

\paragraph{Evaluation metrics} 

To evaluate each generator,
we compare the empirical distributions of real and generated
out-of-sample paths (\cref{subsec:problem_setting}).
We report several discriminative and distributional metrics;
all defined so that lower values indicate better fit.
Discriminative metrics fit binary classifiers to distinguish real from
generated paths and report $|\mathrm{accuracy}-0.5|$.
We use linear classifiers on signature features (\textsc{SIG})
and random MLP features (\textsc{MLP}),
and GRU classifiers applied either to individual paths
(\textsc{RNN}; \citealp{yoon_2019_times})
or sets of paths (\textsc{SRNN}).
Distributional metrics compare autocorrelation (\textsc{ACF}),
cross-correlation (\textsc{CCF}),
marginals (\textsc{CVM}) and their tails (\textsc{ES}),
joint distributions of consecutive observations (\textsc{ED}),
and trading-strategy PnL tails (\textsc{PES}; \citealp{cont_2025_tailg}).
Definitions of all metrics are in \cref{app:subsec:eval_metrics}.

\subsection{Generative models}\label{subsec:gen_models}
\begin{figure*}[t]
    \centering
    \includegraphics[width=\linewidth]{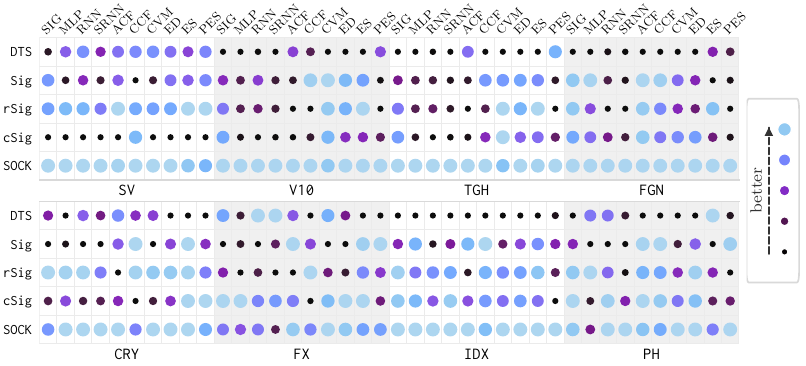}
    \caption{
    Summary of evaluation metrics across selected synthetic
    (\texttt{SV}, \texttt{V10}, \texttt{TGH}, \texttt{FGN})
    and real datasets (\texttt{CRY}, \texttt{FX}, \texttt{IDX}, \texttt{PH}).
    Dot color and size encode the metric value:
    lighter and larger dots indicate a smaller discrepancy between real and generated data
    (hence better).
    }
    \label{fig:grid}
\end{figure*}

We compare several methods for training generative models
and train separate models for each dataset under five random seeds.
On synthetic datasets, each seed resamples the training and evaluation data,
and also changes the training randomness.
On real datasets, we cannot resample the data,
so only the training randomness varies across seeds.

We train three generative models via feature matching on joined past--future
segments (\cref{subsec:nonadversarial_fm})
using three different feature maps:
truncated signature \texttt{Sig} \citep{ni_2022_sigwa},
randomized signature \texttt{rSig} \citep{2024NiklasGononUniversalrandomisedsignaturesgenerative},
and \texttt{SOCK}.
For the random feature maps \texttt{SOCK} and \texttt{rSig},
we resample parameters every $R = 100$ steps;
for \texttt{Sig} the feature map is fixed throughout training.
Further, we benchmark against the Sig-Wasserstein GAN \citep{liao_2024_sigwa},
which uses conditional feature matching with signature features
(\texttt{cSig}).

For the truncated and randomized signature features,
we augment paths before computing the signature,
which significantly improves their performance.
We first apply the Lead-Lag transform,
then augment the path with a time channel,
and finally apply the I-visibility transformation \citep{2020SignatureFeaturesWithTheVisibilityTransformation}.
For \texttt{Sig} and \texttt{cSig}
we use a truncation depth of $3$ \citep{liao_2024_sigwa}
and for \texttt{rSig} we use a hidden dimension of $128$
and the tanh activation function.
For \texttt{SOCK}, we augment with $\mathtt{int}$ and $\mathtt{posneg}$
(Appendix~\ref{app:subsec:sock_augmentations}),
and set $\tau = 0.1$, $M = 256$.
With \texttt{SOCK}, we apply an additional componentwise feature scaling
before computing feature means (Appendix~\ref{app:subsubsec:feature_scaling}).
For \texttt{Sig} and \texttt{rSig},
we found that such a standardization is not beneficial
(\cref{tab:appendix_ablation_feature_scaling}).
We use the above choices on all datasets.

For these four training methods,
we parameterize ${p_\theta(\cdot \mid x^-)}$ with
the same GRU-based generator \citep{chung_2014_empira}.
Given a context $x^-$, we draw $\varepsilon_0 \sim \mathcal{N}(0, I_d)$
and use an MLP on $(x^-,\varepsilon_0)$ to produce the initial GRU state.
We then draw an i.i.d.\ noise sequence
$\varepsilon_{1:T} \sim \mathcal{N}(0, I_d)$
and map it directly to a generated sample path
using a single-layer GRU-based decoder (Appendix~\ref{subsubsec:gen_arch}).

Additionally,
we compare to Diffusion-TS (\texttt{DTS}) \citep{yuan_2024_diffu},
a diffusion model for time series.
Diffusion-TS supports conditional generation by
keeping the conditioning segment $x^-$ fixed
and denoising the masked future segment of length $T$.
We use the authors' implementation and recommended hyperparameters
(Appendix~\ref{app:subsec:baseline_details}).
  
\subsection{Generative Modeling Results}\label{subsec:results}
\Cref{fig:grid} visualizes the evaluation metrics for the generated
distributions on 8 of the 12 datasets.
Lighter, larger dots indicate smaller metric values and thus
smaller discrepancies between real and generated paths.
Full numerical results for all datasets are provided in
Figures~\ref{fig:appendix_heatmap_real_other}-\ref{fig:appendix_heatmap_synth}
of Appendix~\ref{sec:add_results}.

We find that
\textbf{generators trained to match random \texttt{SOCK} features
achieve the strongest overall performance} across datasets,
outperforming the signature baselines and Diffusion-TS on most metrics.
Quantitatively,
\texttt{SOCK} attains the best average rank ($1.49$)
over all dataset--metric pairs, followed by \texttt{rSig} ($2.80$).
On synthetic datasets,
\texttt{SOCK} outperforms all baselines by a wide margin on most metrics.
On real datasets, \texttt{SOCK} also performs best overall,
though not uniformly and with smaller margins.
The smaller gap between methods on real datasets
likely reflects a combination of
typically weak autocorrelation in daily returns, 
noisier evaluation,
and possible distributional drift over the 8-year horizon.
In practice, such drift can be partially mitigated by regularly retraining the generator.
We use the 8-year horizon
to obtain sufficient samples for a stable distributional evaluation.

Because results across datasets and metrics are correlated,
these aggregate ranks should be interpreted with care;
standard significance tests assuming independence are not appropriate here.
Instead, we devote the remainder of this section to analyzing the generative models
at the level of individual metrics and datasets.

\paragraph{Marginal fit}
\begin{figure}[t]
    \centering
    \includegraphics[width=\linewidth]{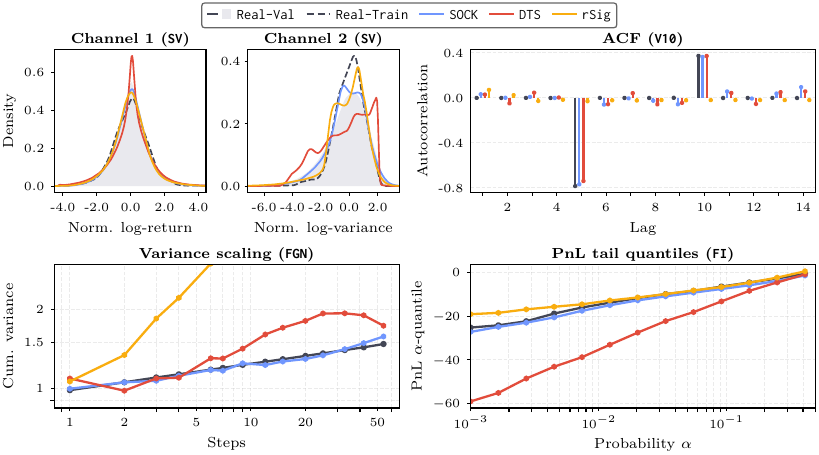}
    \caption{
    Comparison of real and generated distributions;
    visual similarity to real
    distributions indicates high generative fidelity.
    Top-left:  marginal KDEs for the two \texttt{SV} channels.
    Top-right: ACF of the first \texttt{V10} channel;
    the real process has nonzero autocorrelation at every fifth lag.
    Bottom-left: variance of cumulative \texttt{FGN} process $\sum_{i = 1}^\tau X_i$
    (log--log scale); slope on real data is $2 \mathcal{H}$, where $\mathcal H=0.05$.
    Bottom-right: left-tail PnL quantiles from a momentum trading strategy on FI.
    }
    \label{fig:dist_visualizations}
    \vskip -0.10in
\end{figure}

\begin{figure*}[t]
    \centering
    \includegraphics[width=\linewidth]{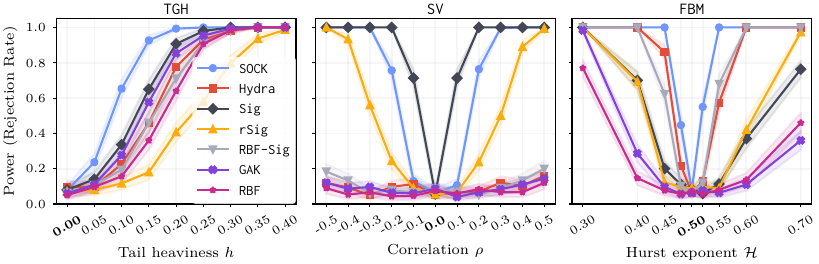}
    \caption{Permutation test power at significance level $\alpha=0.05$ for
    feature- and kernel-based MMDs.
    Bold x-axis values indicate the parameter value under the null hypothesis
    $H_0$.
    Away from $H_0$, higher power means that the MMD more reliably detects the
    parameter change (hence better).}
    \label{fig:hypothesis-testing}
\end{figure*}

\begin{figure}[t]
    \centering
    \includegraphics[width=\linewidth]{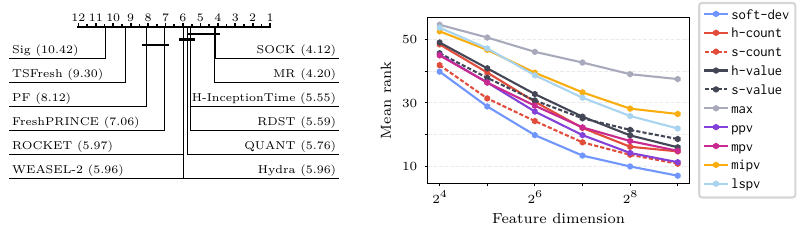}
    \caption{
    Results on UCR classification tasks.
    Left: critical difference diagram for classifiers trained on
    different feature maps.
    Right: mean rank for classifiers using the \texttt{SOCK} backbone
    with different poolings.
    We call Hydra's poolings
    \texttt{h-count} and \texttt{h-value};
    \texttt{s-count} and \texttt{s-value} are differentiable versions of those;
    \texttt{soft-dev} (ours);
    \texttt{max}, \texttt{ppv} (Rocket);
    \texttt{ppv}, \texttt{mpv}, \texttt{mipv}, \texttt{lspv} (MultiRocket).
    }
    \label{fig:classification}
\end{figure}
Across datasets,
\texttt{SOCK} performs best overall on the  marginal-fit metric \textsc{CVM}
(Appendix~\ref{sec:add_results}).
As an example, the top left panel of
\cref{fig:dist_visualizations} shows KDEs
for the two channels of the \texttt{SV} dataset (log-returns and log-variance).
The log-variance marginal differs noticeably in-sample versus out-of-sample,
illustrating how the limited training data
provides a noisy view of the true data distribution.
Despite this gap,
the \texttt{SOCK}-trained generator matches the out-of-sample marginals well
when generating conditionally from out-of-sample contexts $x^{-}_{t}$.
This indicates that the generator captures the dependence structure of the process
and does not merely resample the empirical marginal from the training path
(see also the context-shuffling ablation in \cref{tab:appendix_context_shuffling}).
All methods reproduce the return density reasonably well,
but differences are clearer for log-variance:
\texttt{SOCK} is closest to the out-of-sample marginal,
while \texttt{DTS} deviates substantially.

\paragraph{Autocorrelation}
Among the compared generators, \texttt{SOCK}-trained models
most reliably reproduce autocorrelation structures.
On the oscillatory datasets \texttt{V10} and \texttt{TGH},
only \texttt{SOCK} and \texttt{DTS} manage to reproduce
the distinct autocorrelation pattern
as shown in the top right panel of \cref{fig:dist_visualizations}.
\texttt{SOCK} also best captures the anti-persistent
dependencies of the fractional Gaussian noise process
(bottom left of \cref{fig:dist_visualizations}).
On the real daily-return datasets, autocorrelation is typically weak,
making differences between methods less pronounced.

\paragraph{Tail fit}
For financial applications, it is important that the generator
produces realistic tail-risk scenarios.
Which tail events matter most depends on the precise application.
Following \citet{cont_2025_tailg},
we compare the tails of profit-and-loss (PnL) distributions
obtained by applying simple trading strategies to real and generated data.
Such PnL tails are sensitive to a range of distributional path properties,
including auto- and cross-correlation structure.
The \textsc{PES} metric summarizes such tail discrepancies
(Appendix~\ref{app:subsec:eval_metrics}) and
the bottom right panel of \cref{fig:dist_visualizations}
shows the PnL tail of a momentum strategy on the \texttt{FI} dataset.
Both \texttt{SOCK} and \texttt{rSig} match the PnL tails closely,
while \texttt{DTS} overestimates the tail heaviness.

\paragraph{Ablations}
We study how sensitive generator training by random \texttt{SOCK} feature matching
is to feature-map resampling and to architectural choices in \texttt{SOCK}
(augmentations, kernel width, and poolings).
We run these ablations on the \texttt{TGH} dataset
and report results in Appendix~\ref{app:subsec:ablations}.
We find that resampling the random parameters of the feature map
consistently improves performance for both \texttt{SOCK} and \texttt{rSig}.
The gains from resampling are especially large for the weaker feature map
\texttt{rSig}, while \texttt{SOCK} feature matching already trains good
generative models without resampling.
Regarding \texttt{SOCK}'s architectural choices,
such as augmentations, kernel width, and pooling,
our default choices give the best \textsc{RNN} score,
but the gaps to alternative choices are small
and the ranking varies across metrics.
This shows that random convolutional feature matching
is robust to these choices.
We next benchmark the \texttt{SOCK} feature map
and its architectural choices on purely discriminative tasks.

\section{Discriminative evaluation}\label{sec:discriminative_eval}

\subsection{Hypothesis testing}\label{sec:hypo}

Following prior work \citep{2022SignatureMomentsToCharacterizeLawsOfStochasticProcesses,2021HigherOrderKernelMeanEmbeddings,2023PCFGAN},
we perform a series of two-sample hypothesis tests for stochastic processes, testing the null hypothesis that two stochastic processes, $X_{\beta_0}$ and $X_{\beta_1}$, have the same law. Specifically, we use the maximum mean discrepancy (MMD) \citep{2012MMDAKernelTwoSampleTest} based on either explicit feature vectors or time series kernels, within a permutation test (see \cref{app:subsec:hypotesting}).
Such hypothesis tests serve as a diagnostic for feature matching with the same feature map (or kernel):
if the induced MMD is insensitive to a given mismatch between $X_{\beta_0}$ and $X_{\beta_1}$, then minimizing \cref{eq:feature_matching_loss} cannot correct that mismatch.

We consider three tests that probe different time series characteristics,
using the same parametric processes as in \cref{sec:ts_gen}.
For each process, we vary a single parameter:
\texttt{TGH} (tail heaviness $h$),
\texttt{SV} (correlation $\rho$),
and \texttt{FBM}\footnote{
For consistency with prior work, we use fractional Brownian motion (\texttt{FBM}) rather than \texttt{FGN}.} (Hurst $\mathcal{H}$).
As explicit
feature baselines we include truncated signatures \citep{ni_2022_sigwa}, randomized signatures \citep{2021ExpressivePowerOfRandomizedSignatures}, and Hydra \citep{2023Hydra}.
Kernel baselines include the truncated RBF signature kernel \citep{2019KernelsForSequentiallyOrderedData}, the Global Alignment Kernel (\texttt{GAK}) \citep{2007GlobalAlignmentKernel,2011FastGlobalAlignment}, and the classical RBF kernel. Results are shown in \cref{fig:hypothesis-testing}, with further details in \cref{app:subsec:hypotesting}. \texttt{SOCK} achieves the highest test power on \texttt{FBM} and \texttt{TGH}, and ranks second on \texttt{SV}.
Notably, \texttt{SOCK} performs well on \texttt{SV}, a task where most baselines, including Hydra, barely outperform random guessing.
Additionally, \texttt{SOCK} is robust to hyperparameter tuning, in comparison to signatures which are very sensitive to hyperparameters. 

\subsection{Time series classification}\label{subsec:classification}
To further assess the expressivity of \texttt{SOCK} features, we evaluate them on the 112 univariate time series classification datasets of the UCR repository \citep{2024BakeOff}.
Following prior works \citep{2020Rocket, 2022MultiRocket, 2023Hydra},
we fix the classification pipeline to a feature extraction step followed by a RidgeCV classifier.
First, we compare the poolings from Rocket, MultiRocket, and Hydra
to three differentiable poolings:
soft-deviation, the \texttt{SOCK} default,
and soft-count and soft-value, differentiable generalizations of
Hydra's count and value poolings
(Appendix~\ref{app:subsec:sock_pooling}).
This is done by keeping the preprocessing pipeline and random kernels fixed,
while varying the temporal pooling operator.
The right panel of \cref{fig:classification} shows that classifiers trained with
soft-deviation features consistently outperform other popular poolings across feature dimensions.
Comprehensive experimental details, including setup, baselines, and robustness ablations (on random bias, temperature $\tau$, number of kernels $k$), are provided in Appendix~\ref{app:subsec:tsc}.

Second, we compare \texttt{SOCK} against state-of-the-art models identified in \citet{2024BakeOff}
via a critical difference diagram in the left panel of \cref{fig:classification}.
We find that \texttt{SOCK} achieves the best average rank among the compared models.
\texttt{SOCK} achieves this using an order of magnitude fewer random features
and only a single pooling operation, compared to MultiRocket's four
(details in Appendix~\ref{app:subsec:tsc}).
These results show that \texttt{SOCK} performs strongly in purely
discriminative settings
and support its use as a feature map for generative modeling.

\section{Conclusions and limitations}\label{sec:conclusions_limitations}

In this work, we studied time series generation in a limited-data setting
relevant to many practical use cases in finance.
We showed that conditional generative models
trained by random convolutional feature matching achieve high generative fidelity,
and outperform signature-based feature matching and Diffusion-TS on most metrics
across small-sample synthetic and real financial datasets.
To enable this approach, we introduced \texttt{SOCK},
a differentiable random convolutional feature map
based on soft kernel competitions.
We showed that \texttt{SOCK} also performs well on discriminative tasks,
matching or outperforming existing random convolutional feature maps.

Motivated by our target applications,
we focus on low-dimensional datasets with a single historical path.
While random convolutional feature matching can be applied to larger datasets,
we do not study that regime.
Learned discriminators and diffusion models are likely to benefit more from
additional data than random feature maps.
Moreover, we use a deliberately simple generator architecture
for all feature matching methods.
Stronger architectures could improve performance independently of feature-map
improvements.
Finally, the strong results on the UCR datasets
motivate future work on \texttt{SOCK} for time series classification.

\section*{Acknowledgements}
KM is supported by JPMorgan Chase \& Co. through the EPSRC Centre for Doctoral Training in Mathematics of Random Systems: Analysis, Modelling and Simulation (EPSRC Grant EP/S023925/1).
NZ has been supported by the Roth Scholarship at Imperial College London. The work of TC was supported in part by UK Research and Innovation (UKRI) through the Engineering and Physical Sciences Research Council (EPSRC) via Programme Grants [Grant No. UKRI1010: High order mathematical and computational infrastructure for streamed data that enhance contemporary generative and large language models]. We acknowledge computational resources and support provided by the Imperial College Research Computing Service (DOI: \texttt{10.14469/hpc/2232}). For the purpose of open access, the authors have applied a Creative Commons Attribution (CC BY) licence to any Author Accepted Manuscript version arising.

A preliminary version of this work appeared at the NeurIPS 2025
Workshop on Generative AI in Finance
\citep{2025GeneratingTimeSeriesByMatchingRandomConvolutionalFeatures}.

{
\small
\bibliographystyle{plainnat}
\bibliography{references,rconv}
}

\clearpage
\appendix

\AppendixToC
\clearpage

\section{Detailed results and ablations for generative experiments} \label{sec:add_results}

\subsection{Detailed numerical results}

\begin{figure*}[!h]
    \centering
    \includegraphics[width=\linewidth]{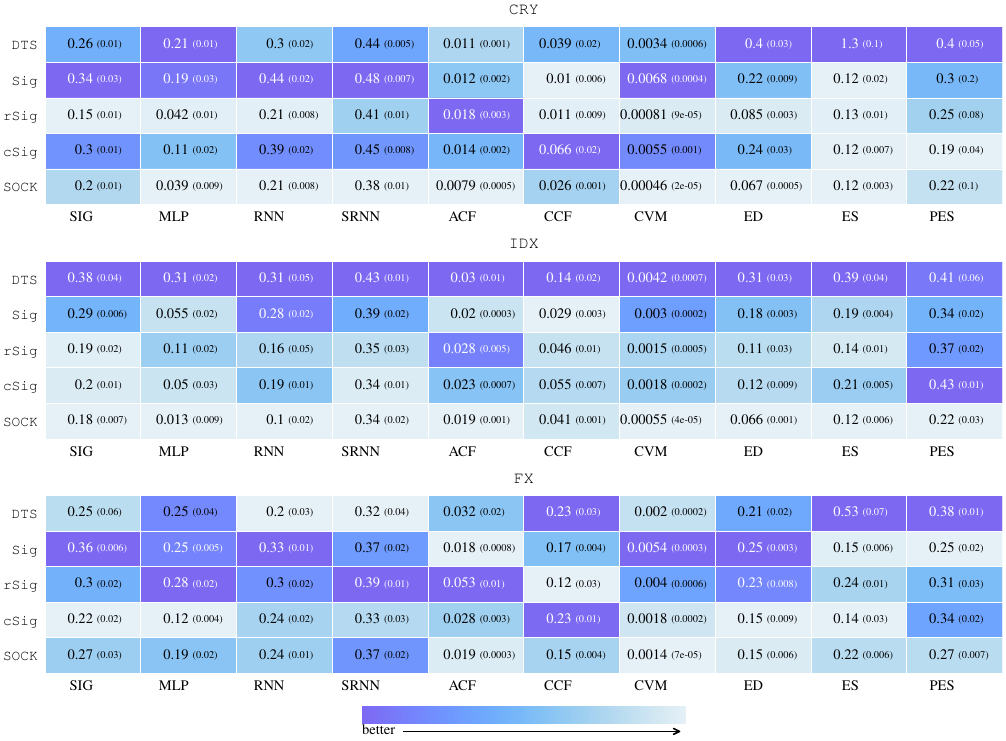}
    \caption{
    Detailed results on the real datasets \texttt{CRY}, \texttt{IDX}, and \texttt{FX}.
    Each cell reports the mean metric value across 5 seeds (standard deviation in parentheses).
    Here, seeds vary only the training randomness. Cell color encodes the mean value (lighter is better).
    The color is based on the averaged metrics.}
    \label{fig:appendix_heatmap_real_other}
\end{figure*}

\begin{figure*}[!h]
    \centering
    \includegraphics[width=\linewidth]{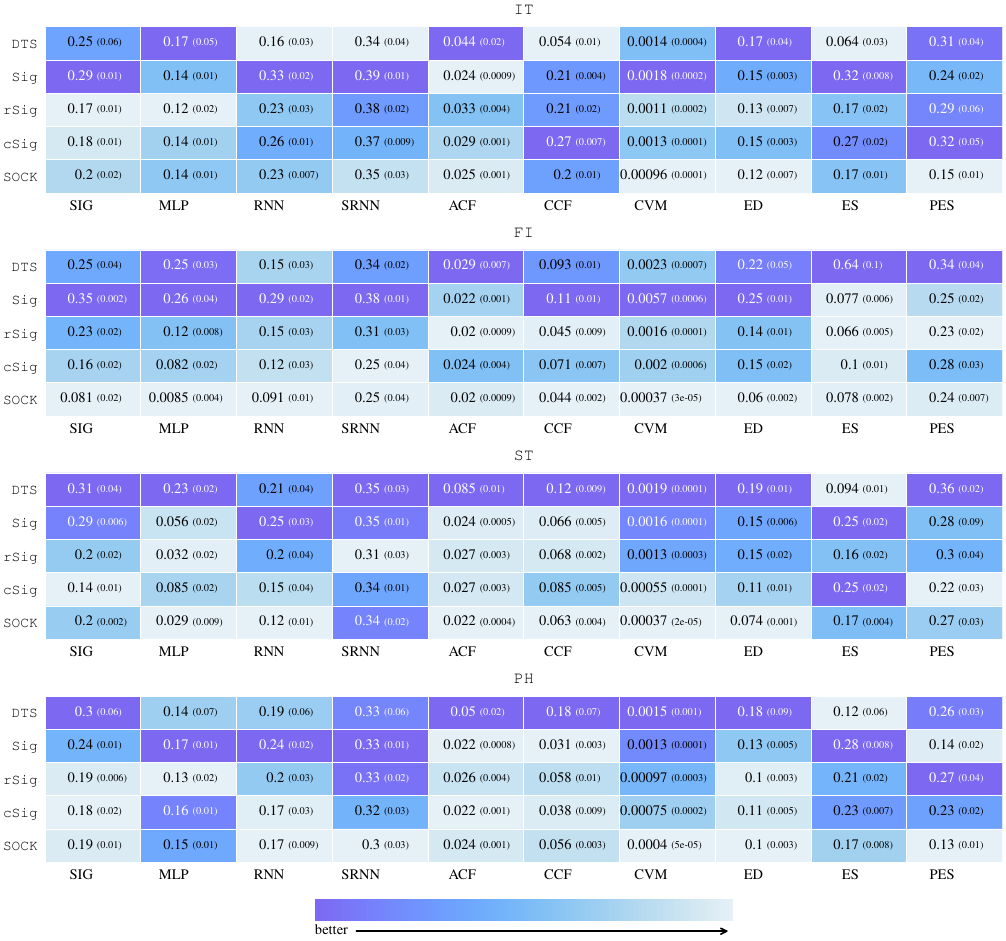}
    \caption{
    Detailed results on the real baskets-of-stocks datasets.
    Each cell reports the mean metric value across 5 seeds (standard deviation in parentheses).
    Here, seeds vary only the training randomness. Cell color encodes the mean value (lighter is better).
    The color is based on the averaged metrics.}
    \label{fig:appendix_heatmap_real_stocks}
\end{figure*}

\begin{figure}[!h]
    \centering
    \includegraphics[width=\linewidth]{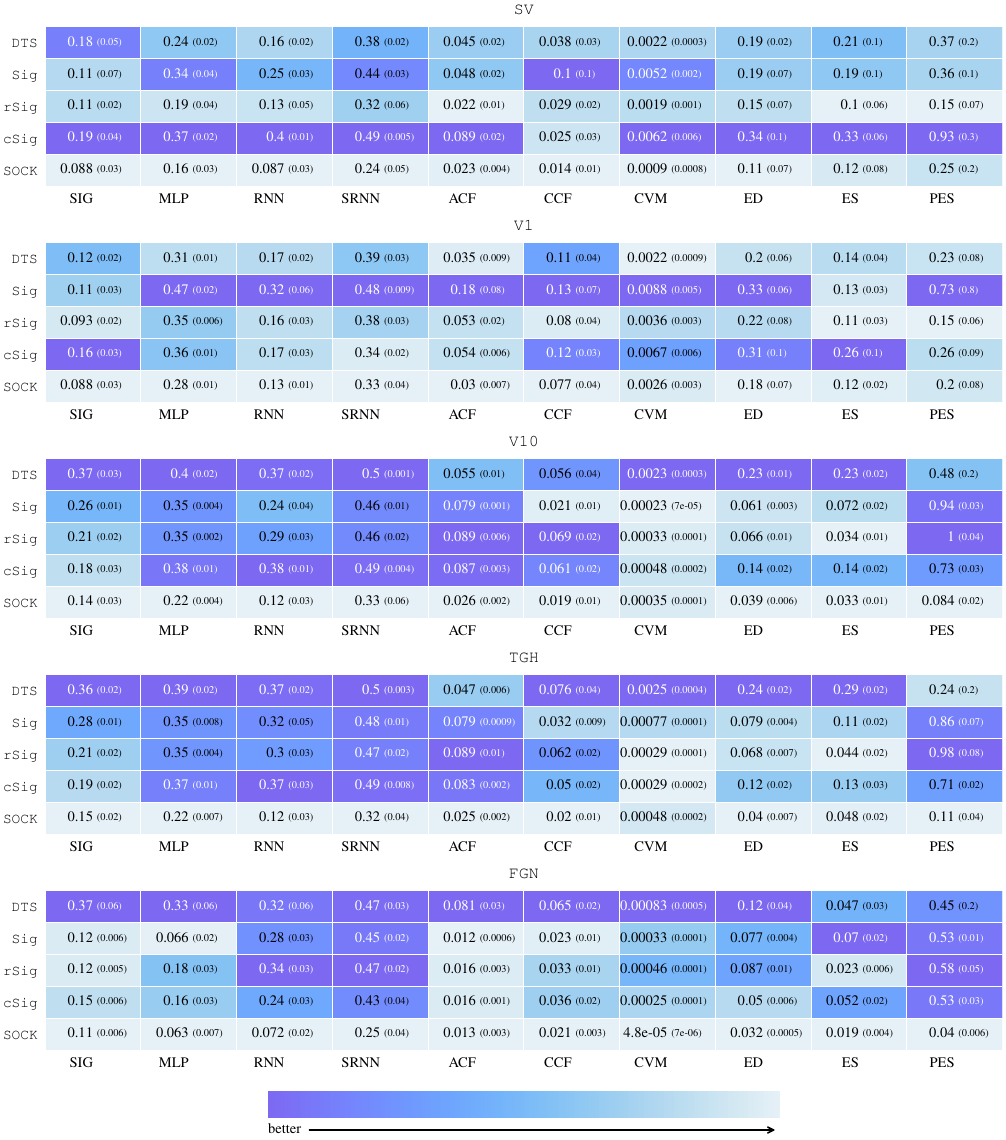}
    \caption{
    Detailed results on synthetic datasets.
    Each cell reports the mean metric value across 5 seeds (standard deviation in parentheses), where seeds vary both the data draw and training randomness. Cell color encodes the mean value (lighter is better).
    The color is based on the averaged metrics.
    }
    \label{fig:appendix_heatmap_synth}
\end{figure}

\clearpage

\subsection{Ablations}\label{app:subsec:ablations}

\begin{table}[h]
\caption{
Ablations for the generative experiments.
All ablations are run on the \texttt{TGH} dataset and repeated over five random seeds.
Reported metrics are averaged over seeds; lower is better, and bold marks the best value within each block.
Metric definitions are given in Appendix~\ref{app:subsec:eval_metrics}.
We report \textsc{RNN} and distributional metrics.}
\label{tab:appendix_ablation_abs_combined}
\begin{center}
\begingroup
\scshape
\begin{small}
\begin{tabular}{lrrrrrr}
\toprule
\multicolumn{7}{l}{\textbf{Ablation 1: Resample}} \\
Method & \makebox[2.0em]{RNN} & \makebox[2.0em]{ACF} & \makebox[2.0em]{CCF} & \makebox[2.0em]{CVM} & \makebox[2.0em]{ED} & \makebox[2.0em]{ES} \\
\midrule
\texttt{SOCK} (resample $\times$) & 0.126 & 0.027 & \textbf{0.019} & 0.00145 & 0.041 & 0.051 \\
\texttt{SOCK} (resample \checkmark) & \textbf{0.117} & \textbf{0.025} & 0.020 & \textbf{0.00048} & \textbf{0.040} & \textbf{0.048} \\
\cmidrule(lr){1-7}
\texttt{rSig} (resample $\times$) & 0.416 & 0.093 & 0.072 & 0.00249 & 0.172 & 0.080 \\
\texttt{rSig} (resample \checkmark) & \textbf{0.296} & \textbf{0.089} & \textbf{0.062} & \textbf{0.00029} & \textbf{0.068} & \textbf{0.044} \\
\midrule
\addlinespace[1em]
\multicolumn{7}{l}{\textbf{Ablation 2: Augmentations}} \\
Augmentations & \makebox[2.0em]{RNN} & \makebox[2.0em]{ACF} & \makebox[2.0em]{CCF} & \makebox[2.0em]{CVM} & \makebox[2.0em]{ED} & \makebox[2.0em]{ES} \\
\midrule
$\mathcal{A}(X) = X$ & 0.127 & \textbf{0.021} & \textbf{0.018} & 0.00062 & 0.042 & 0.054 \\
$\mathcal{A}(X) = \mathtt{int}(X)$ & 0.130 & 0.023 & \textbf{0.018} & \textbf{0.00043} & 0.042 & 0.066 \\
$\mathcal{A}(X) = \mathtt{posneg}(\mathtt{int}(X))$ & \textbf{0.117} & 0.025 & 0.020 & 0.00048 & \textbf{0.040} & \textbf{0.048} \\
\midrule
\addlinespace[1em]
\multicolumn{7}{l}{\textbf{Ablation 3: Kernel width}} \\
Kernel width W & \makebox[2.0em]{RNN} & \makebox[2.0em]{ACF} & \makebox[2.0em]{CCF} & \makebox[2.0em]{CVM} & \makebox[2.0em]{ED} & \makebox[2.0em]{ES} \\
\midrule
W=1 & 0.119 & \textbf{0.022} & \textbf{0.017} & \textbf{0.00046} & 0.042 & 0.050 \\
W=2 & \textbf{0.117} & 0.025 & 0.020 & 0.00048 & \textbf{0.040} & \textbf{0.048} \\
\midrule
\addlinespace[1em]
\multicolumn{7}{l}{\textbf{Ablation 4: Pooling}} \\
Pooling & \makebox[2.0em]{RNN} & \makebox[2.0em]{ACF} & \makebox[2.0em]{CCF} & \makebox[2.0em]{CVM} & \makebox[2.0em]{ED} & \makebox[2.0em]{ES} \\
\midrule
soft-dev & \textbf{0.117} & 0.025 & \textbf{0.020} & 0.00048 & \textbf{0.040} & 0.048 \\
soft-count & 0.132 & \textbf{0.024} & 0.024 & \textbf{0.00027} & 0.041 & 0.048 \\
soft-value & 0.119 & \textbf{0.024} & 0.021 & 0.00050 & \textbf{0.040} & \textbf{0.043} \\
\bottomrule
\end{tabular}
\end{small}
\endgroup
\end{center}
\vspace{2em}
\emph{Remarks.}
Ablation 1 compares training with a fixed random feature map against resampling
the feature map every $R=100$ steps; resampling improves both \texttt{SOCK} and \texttt{rSig}, with larger gains for \texttt{rSig}.
Ablation 2 varies the input augmentation $\mathcal{A}$, defined in
Appendix~\ref{app:subsec:sock_augmentations}; adding \texttt{posneg}
improves \textsc{RNN}, but weakens some other metrics.
Ablation 3 varies the kernel width $W$ of the grouped random convolutions;
both settings give good results, with $W=2$ best on \textsc{RNN}.
Ablation 4 varies the pooling applied to the soft kernel competitions, with
poolings defined in Appendix~\ref{app:subsec:sock_pooling}; all three poolings are competitive, with \texttt{soft-dev} best on \textsc{RNN}.

\medskip
\emph{Limitations.}
These ablations are limited to the \texttt{TGH} dataset,
which has a complex autocorrelation structure and heavy-tailed and skewed marginals,
but its cross-correlation structure is rather simple.
Results may therefore look different on other datasets.

\medskip
\emph{Additional ablations.}
We additionally study choices of the \texttt{SOCK} architecture
on purely discriminative tasks in \cref{sec:discriminative_eval}
and Appendix~\ref{app:subsec:tsc}.
These also allow us to benchmark \texttt{SOCK} against other random convolutional
feature maps that are non-differentiable and that therefore
cannot be studied in the generative setting.

\end{table}

\newpage
\subsection{Additional visualizations} \label{app:sec:add_visualizations}

\begin{figure}[!h]
    \centering
    \includegraphics[width=\linewidth]{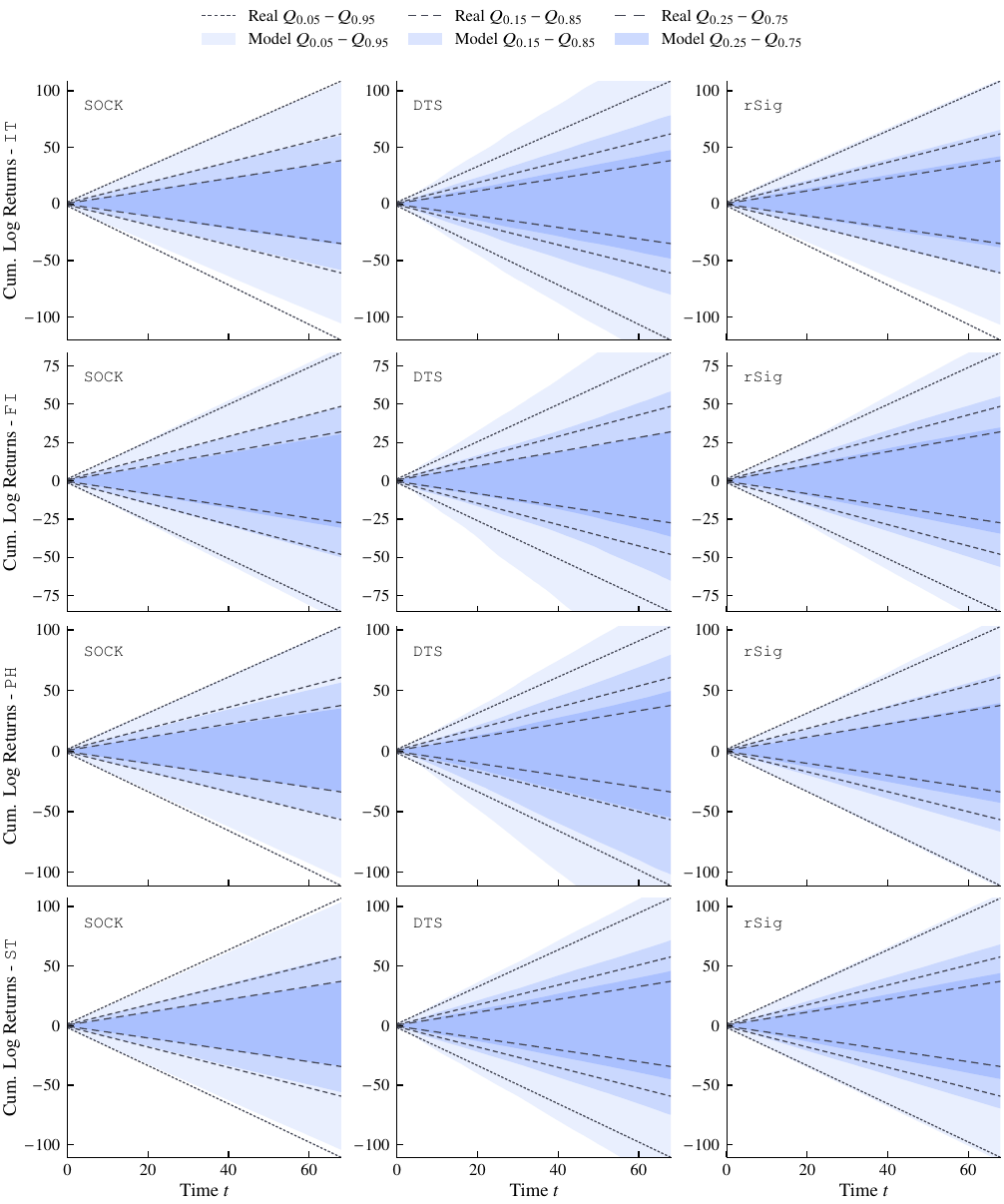}
    \caption{
    Real vs.\ model quantile bands over time for cumulative log returns
    on the first channel of each stock dataset.
    \texttt{SOCK} and \texttt{rSig} align more closely with the real quantiles than \texttt{DTS}.
    }
\end{figure}

\clearpage
\section{Additional details and pseudocode for \texttt{SOCK}}\label{sec:code}
This appendix provides implementation details for the \texttt{SOCK} feature map
from \cref{sec:sock}.
We discuss our implementation choices and
define alternative choices used in our ablations.

\subsection{Augmentations and normalization}\label{app:subsec:sock_augmentations}

In the first step of \texttt{SOCK}, we compute
\begin{align*}
Y_t = P\,\mathcal{N}(\mathcal{A}(X))_t ,
\end{align*}
where $\mathcal{A}$ is an augmentation map,
$\mathcal{N}$ is a normalization function,
and $P$ is a random projection matrix.
In this section, we provide details on the choice of $\mathcal{A}$ and $\mathcal{N}$.

\paragraph{Augmentations}
We make different augmentation choices between
our generative and discriminative experiments.
We provide an ablation in Appendix~\ref{app:subsec:ablations} and use this section to define all the choices considered.
When applying \texttt{SOCK} to a new task we recommend trying
either $\mathcal{A}(X) = \mathtt{int}(X)$ (when $X$ is stationary)
or $\mathcal{A}(X) = \mathtt{diff}(X)$ (when $X$ is non-stationary).

Let $U\in\mathbb{R}^{T\times d_U}$ be an input path.
We use augmentation primitives that preserve $U$ and append derived channels:
\begin{align*}
    \mathtt{diff}(U)_t
    &=
    [U_t,\; \Delta U_t], \\
    \mathtt{int}(U)_t
    &=
    [U_t,\; {\textstyle\sum_{s=1}^t} U_s],
    \\
    \mathtt{posneg}(U)_t
    &=
    [U_t,\; U_t^+,\; U_t^-].
\end{align*}
Here $\Delta U_1=0$, $\Delta U_t=U_t-U_{t-1}$ for $t>1$,
$U_t^+=\max(U_t,0)$, and $U_t^-=\min(U_t,0)$,
with maximum and minimum applied elementwise.

In the generative experiments, we use
\begin{equation*}
    \mathcal{A}(X)=\mathtt{posneg}(\mathtt{int}(X)) .
\end{equation*}
We compare this choice to no augmentation, $\mathcal{A}(X)=X$,
and to $\mathcal{A}(X)=\mathtt{int}(X)$ in Appendix~\ref{app:subsec:ablations}.

\paragraph{Normalization}
For an augmented path $U=\mathcal{A}(X)$, the normalization $\mathcal{N}$
centers and scales each channel using statistics fitted on the augmented
training paths:
\begin{equation*}
    \mathcal{N}(U)_t
    =
    (U_t-\mu)\odot\sigma^{-1},
\end{equation*}
where $\mu,\sigma\in\mathbb{R}^{d'}$ are computed over all training paths
and time indices, and $\odot$ denotes elementwise multiplication.

\subsection{Random parameters and dilations}\label{app:subsec:sock_random}
The random parameters of \texttt{SOCK} consist of the projection matrix $P$
and the convolutional kernels for each dilation.
We draw a pre-projection matrix
$\hat P\in\mathbb{R}^{M\times d'}$ with i.i.d.\ $N(0,1)$ entries
and normalize each row:
\begin{equation*}
    P_{i,\cdot}
    =
    \frac{\hat P_{i,\cdot}}{\lVert \hat P_{i,\cdot}\rVert_2}.
\end{equation*}
We choose $M$ divisible by the group width $W$ and set $G=M/W$.
For each dilation $\delta$, group $g$, and kernel index $k$,
we draw a pre-kernel
$\hat{w}^{(g,k)}\in\mathbb{R}^{W\times L}$
with i.i.d.\ $N(0,1)$ entries, suppressing $\delta$ in the notation.
We then center and normalize:
\begin{equation*}
    w^{(g,k)}
    =
    \frac{
        \hat w^{(g,k)}
        -
        \mathrm{mean}(\hat w^{(g,k)})
    }{
        \lVert
        \mathrm{vec}\!\left(
        \hat w^{(g,k)}
        -
        \mathrm{mean}(\hat w^{(g,k)})
        \right)
        \rVert_1
    } .
\end{equation*}
Thus each kernel has zero sum and unit $\ell_1$ norm.
This scheme is a straightforward extension of Hydra’s \citep{2023Hydra}
kernel sampling to $W >1$ and
ensures that, a priori, each kernel
has the same chance at winning the local competition.

We follow Hydra's dilation scheme \citep{2023Hydra}.
For input length $n$ and kernel length $L=9$,
\begin{equation*}
    \mathcal{D}
    =
    \{2^e:\ e=0,1,\dots,e_{\max}\},
    \qquad
    e_{\max}
    =
    \left\lfloor
    \log_2\!\left(\frac{n-1}{L-1}\right)
    \right\rfloor .
\end{equation*}
For segments of length $n=q+T = 69$, this gives $\mathcal{D} = \{1,2,4,8\}$.

\subsection{Pooling operators}\label{app:subsec:sock_pooling}
For each group $g$ and time index $t$,
the $K$ convolution responses define a competition between kernels.

For $z\in\mathbb{R}^K$, let
\begin{equation*}
    \pi_\tau(z)_k
    =
    \frac{\exp(z_k/\tau)}{\sum_{j=1}^{K}\exp(z_j/\tau)}
\end{equation*}
denote the $\mathrm{softmax}$ at temperature $\tau > 0$.

Writing $Z_t^{(g,\cdot)}=(Z_t^{(g,1)},\dots,Z_t^{(g,K)})$,
we define the following differentiable pooling operators:
\begin{alignat*}{2}
    F_{\mathrm{soft\text{-}dev}}^{(g,k)}
    &=
    \bigl[
    \frac{1}{T}
    \sum_{t=1}^{T}
    \bigl(
    P_t^{(g,k)}
    -
    \frac{1}{T}\sum_{s=1}^{T}P_s^{(g,k)}
    \bigr)^2
    \bigr]^{1/2},
    \qquad&
    P_t^{(g,k)}
    &=
    \pi_\tau(Z_t^{(g,\cdot)})_k,
    \\
    F_{\mathrm{soft\text{-}count}}^{(g,k)}
    &=
    \frac{1}{T}
    \sum_{t=1}^{T}
    P_t^{(g,k)},
    \qquad&
    P_t^{(g,k)}
    &=
    \pi_\tau(Z_t^{(g,\cdot)})_k,
    \\
    F_{\mathrm{soft\text{-}value}}^{(g,k)}
    &=
    \frac{1}{T}
    \sum_{t=1}^{T}
    \tilde P_t^{(g,k)} Z_t^{(g,k)},
    \qquad&
    \tilde P_t^{(g,k)}
    &=
    \pi_\tau(-Z_t^{(g,\cdot)})_k .
\end{alignat*}

The first statistic is the softmax-deviation pooling, which is \texttt{SOCK}'s default pooling.
The other two are differentiable analogues of Hydra's count and value features;
the value feature uses $\pi_\tau(-Z_t^{(g,\cdot)})$ to match Hydra's
minimum-response convention.

\newpage

\subsection{Pseudocode}

\begin{tcolorbox}[codebox]
\begin{minted}[breaklines,autogobble,fontsize=\scriptsize]{python}
import math
import torch
from torch import Tensor, nn

# the Aug modules map an input tensor x to an augmented version:
#      (B, n_steps, n_channels) -> (B, n_steps, n_channels + n_add_channels)
AUGMENTATIONS: dict[nn.Module] = {"cumsum": CumSumAug, "diff": DiffAug, "posneg": PosNegAug}

def normalize_kernel_(w: Tensor, eps: float = 1e-6) -> None:
    w.sub_(w.mean(dim=(-2, -1), keepdim=True)) # dim -2 is kernel width and dim -1 is kernel length
    w.div_(w.abs().sum(dim=(-2, -1), keepdim=True).clamp_min(eps))

def normalize_proj_(w: torch.Tensor, eps: float = 1e-6) -> None:
    w.div_(w.norm(p=2, dim=1, keepdim=True).clamp_min(eps))

class SOCK(nn.Module):
    def __init__(self,
        n_steps: int,  # corresponds to T in the paper
        n_channels: int,  # corresponds to d in the paper
        tau: float = 0.1,  # softmax temperature
        k: int = 8,  # number of kernels per group
        mix_dim: int = 256,  # corresponds to M in the paper
        kernel_len: int = 9,
        augs: tuple[str, ...] = ("cumsum",),
    ) -> None:
        super().__init__()
        self.tau, self.k = tau, k

        self.augs = nn.Sequential(*[AUGMENTATIONS[aug]() for aug in augs])
        n_channels += sum(aug.n_add_channels for aug in self.augs)
        self.proj = nn.Linear(n_channels, mix_dim, bias=False)

        emax = math.log2((n_steps - 1) / (kernel_len - 1))
        self.dilations = (2 ** torch.arange(int(emax) + 1)).int()
        kernel_width = 2
        self.convs = nn.ModuleList()
        for d in self.dilations:
            self.convs.append(
                nn.Conv1d(
                    in_channels=mix_dim,
                    out_channels=k * (mix_dim // kernel_width),
                    kernel_size=kernel_len,
                    padding="same",
                    dilation=d,
                    groups=mix_dim // kernel_width,
                    bias=False,
                )
            )

        for p in self.parameters():  # SOCK's parameters are untrained, only need gradient w.r.t input x
            p.requires_grad = False
        self.resample()

    def resample(self) -> None:
        self.proj.weight.normal_()
        normalize_proj_(self.proj.weight)
        for conv in self.convs:
            conv.weight.normal_()
            normalize_kernel_(conv.weight)

    def fit_input_scales(self, x: Tensor) -> None: ...  # fits (input_mean, input_scl)
    def fit_ft_scales(self, x: Tensor) -> None: ...  # fits (ft_mean, ft_scl); call after every resample

    def pool(self, z: Tensor) -> Tensor:  # soft-deviation pooling
        return torch.std(torch.softmax(z / self.tau, dim=2), dim=-1)

    def forward(self, x: Tensor) -> Tensor:  # x.shape = (B, n_steps, n_channels)
        x = self.augs(x)
        x = (x - self.input_mean) / self.input_scl
        x = self.proj(x).permute(0, 2, 1)  # (B, mix_dim, n_steps)

        feats = []
        for conv in self.convs:  # loop over dilations; for T=64, use dilations 1,2,4
            z = conv(x)  # (B, n_groups * k, n_steps), where n_groups = mix_dim // kernel_width
            z = z.view(x.size(0), -1, self.k, x.size(-1))  # (B, n_groups, k, n_steps)
            f = self.pool(z)  # (B, n_groups, k)
            feats.append(f.view(x.size(0), -1))  # (B, n_groups * k)
        feats = torch.cat(feats, dim=-1)  # (B, n_dilations * n_groups * k)
        return (feats - self.ft_mean) / self.ft_scl
\end{minted}
\end{tcolorbox}

\clearpage
\section{Additional details on time series generation experiments}\label{sec:additional_details}

\subsection{Datasets}\label{app:subsec:data}

\textbf{Vector Autoregressive Process - \texttt{V1}}.
The process \texttt{V1} is a $3$-dimensional vector autoregressive process
\begin{equation*}
     X_t = \phi X_{t-1} + \varepsilon_t, \quad
    \varepsilon_t \sim N(0, \Sigma),
\end{equation*}
where
$\Sigma = \tfrac{1}{2}J + \tfrac{1}{2}I$ \citep{liao_2024_sigwa}
and $J$ is a matrix where each entry is equal to $1$.
We set $\phi = 0.99$, meaning the process exhibits strong autocorrelation,
which should be simple to detect and generate.
We include it as a first benchmark,
following prior work
\citep{liao_2024_sigwa,2024NiklasGononUniversalrandomisedsignaturesgenerative}.

\textbf{Vector Autoregressive Process - \texttt{V10}}.
The process \texttt{V10} is a $3$-dimensional vector autoregressive process
with oscillating autocorrelation
presented in \citet{brockwell_2016_intro},
where
\begin{equation}\label{eq:var10}
     X_t = \phi_1 X_{t-5} + \phi_2 X_{t-10} + \varepsilon_t, \quad
    \varepsilon_t \sim N(0, \Sigma),
\end{equation}
with coefficients
$\phi_1 = 2r\cos(\omega \pi)$
and
$\phi_2 = - r^2$.
We set $r = \omega = 0.8$
and use innovation covariance $\Sigma = \tfrac{1}{15}(\tfrac{1}{2}J + \tfrac{1}{2}I)$,
which has trace $0.2$.
With this process, we can test the generative model's
ability to reproduce complex autocorrelation patterns.

\textbf{Tukey $g$ and $h$ - \texttt{TGH}}.
The process \texttt{TGH} is an instance of the class of processes
studied by \citet{yan_2019_nonga}.
The process is generated by first simulating
the \texttt{V10} process from \cref{eq:var10}
and then applying pointwise the 
Tukey $g$ and $h$ transform \citep{tukey_1977_explo},
which is defined as
\begin{equation*}
\tau_{g,h}(x)\;=\;
\begin{cases}
\dfrac{e^{g x}-1}{g}\,\exp\!\left(\dfrac{h x^{2}}{2}\right), & g\neq 0,\\
x\,\exp\!\left(\dfrac{h x^{2}}{2}\right), & g=0 .
\end{cases}
\end{equation*}
We set $g=-0.2$ and $h=0.2$ to induce a negative skew
and heavy tails on the marginals of the process.
We apply the transform at the Gaussian process's simulation scale,
then center and scale each transformed channel using statistics fitted on the training path.

\textbf{Stochastic volatility - \texttt{SV}}.
The process \texttt{SV} is a $2$-dimensional process
based on the stochastic volatility model by \citet{heston_1993_a}:
\begin{align*}
    d S_t &= S_t \sqrt{V_t} dW_t ,\\
    d V_t &= \kappa(\theta - V_t)dt + \xi \sqrt{V_t} d B_t .
\end{align*}
We set
$\kappa = 1.0$, $\theta = 0.04$, $\xi = 0.25$, $\rho = -0.7$,
and simulate $(S, V)$ paths using the scheme of \citet{broadie_2006_exact}
at a time discretization of $\Delta t = 1  / 250$.
We then transform $(S, V)$ into
\begin{equation*}
    X_t := (\log S_t - \log S_{t - \Delta t}, \; \log V_t) ,
\end{equation*}
which is a stationary Markov process.
We include this process in our study due to
its practical relevancy in finance
and because accurately generating the process
requires correctly reproducing the spot-vol correlation $\rho$,
as well as the skewed marginals of $\log V$.

\textbf{Fractional Gaussian Noise - \texttt{FGN}}.
The process \texttt{FGN} is a $3$-dimensional Gaussian process with independent channels.
For each $k\in\{1,2,3\}$, let $(B^{\mathcal{H}_k}_t)_{t\in\mathbb{R}}$ be a fractional Brownian motion
with Hurst parameter $\mathcal{H}_k$ \citep{mandelbrot_1968_fract}.
We generate sequences of length $T=64$ by discretizing the time interval $[0,1]$
using the uniform grid ($t_i = i/T$),
and defining the $k$-th channel as the increment process
\begin{equation*}
    X^{(k)}_i \;:=\; B^{\mathcal{H}_k}_{t_i} - B^{\mathcal{H}_k}_{t_{i-1}},
    \qquad i=1,\dots,T.
\end{equation*}
We set $X_i := (X^{(1)}_i, X^{(2)}_i, X^{(3)}_i)\in\mathbb{R}^3$.
We use distinct Hurst parameters
$\mathcal{H}_1 = 0.05$, $\mathcal{H}_2 = 0.15$, and $\mathcal{H}_3 = 0.25$,
so each channel exhibits antipersistent, slowly decaying correlations,
making the temporal dependence non-Markovian and challenging to reproduce.

\textbf{Real datasets (daily)}.
We construct six daily datasets from Yahoo Finance.\footnote{
\url{https://finance.yahoo.com/}.
Yahoo Finance data are subject to Yahoo's terms of use and are not redistributed.
}
We download daily close prices over 2009--2025,
computing log-returns $r_t=\log(P_t/P_{t-1})$, and splitting the resulting series into a training period (2009--2017) and a validation period (2017--2025).
The datasets and their tickers are listed in \cref{tab:real_datasets_daily}.

\begin{table}[t]
\centering
\caption{Daily real datasets}
\label{tab:real_datasets_daily}
\small
\begin{tabular}{@{}lcl@{}}
\toprule
Dataset & $d$ & Tickers \\
\midrule
\texttt{FI}  & 4 & \texttt{JPM}, \texttt{DB}, \texttt{UBS}, \texttt{HSBC} \\
\texttt{PH}  & 4 & \texttt{PFE}, \texttt{JNJ}, \texttt{AZN}, \texttt{NVS} \\
\texttt{IT}  & 4 & \texttt{MSFT}, \texttt{GOOGL}, \texttt{NVDA}, \texttt{AAPL} \\
\texttt{ST}  & 4 & \texttt{PG}, \texttt{NSRGY}, \texttt{KO}, \texttt{UL} \\
\texttt{IDX} & 3 & \texttt{\string^GSPC}, \texttt{\string^VIX}, \texttt{GC=F} \\
\texttt{FX}  & 3 & \texttt{EURUSD=X}, \texttt{JPYUSD=X}, \texttt{GBPUSD=X} \\
\bottomrule
\end{tabular}
\end{table}

\textbf{Cryptocurrency dataset (\texttt{CRY}, intraday)}.
The dataset \texttt{CRY} is constructed from Binance public market data.\footnote{
\url{https://data.binance.vision/}.
The accompanying Binance public-data repository is MIT licensed:
\url{https://github.com/binance/binance-public-data}.
}
We use BTC/USD and ETH/USD close prices,
extract the period 2025-01-01 to 2025-03-01,
resample to 5-minute frequency by last price,
compute log-returns,
and split into a training period (2025-01-01 to 2025-02-01)
and a validation period (2025-02-01 to 2025-03-01).

\paragraph{Higher-dimensional datasets.}
Note that all our experiments focus on time series with a moderate number of channels.
This is a deliberate choice since learning a flexible conditional simulator
from a single historical path is already statistically challenging,
and increasing the channel dimension without more data
makes the problem more underdetermined.
High-dimensional return generation is commonly studied under different
simplifying assumptions:
\citet{2025DiffusionFactorModels} impose a low-dimensional factor structure,
while \citet{2025FactorBasedConditionalDiffusionModel} generate next-day returns
conditional on asset-specific factors.
These settings differ from generating future paths conditional on a recent observed path.
Motivated by these works, our generative modelling approach can be extended
to higher-dimensional financial data by estimating a low-dimensional factor representation,
for example $X_t \approx B F_t + \varepsilon_t$ with $F_t \in \mathbb{R}^k$ and $k \ll d$,
and applying random convolutional feature matching to the factor process.
We leave this extension to future work.

\subsection{Evaluation}\label{app:subsec:eval_metrics}

\subsubsection{Evaluation protocol}

We evaluate generators by comparing empirical distributions of joined
past--future segments, as defined in \cref{subsec:problem_setting}.
For each evaluation time $t$, we compare the realized segment
$x_t^- \oplus x_t^+$ to the generated segment
$x_t^- \oplus \hat x_t^+$, where both segments share the same
length-$q$ context $x_t^-$ and differ only in their length-$T$ continuation.

On real datasets, we use evaluation times
$\mathcal{T}_{\mathrm{eval}}=\{H,H+1,\dots,H+N-T\}$ along the single
out-of-sample trajectory.
Using overlapping segments is necessary to obtain
enough evaluation samples,
but reduces the effective sample size.
The resulting empirical collections are
\begin{equation*}
\{x_t^- \oplus x_t^+\}_{t\in\mathcal{T}_{\mathrm{eval}}}
\qquad\text{and}\qquad
\{x_t^- \oplus \hat x_t^+\}_{t\in\mathcal{T}_{\mathrm{eval}}}.
\end{equation*}
On synthetic datasets, we sample $J$ independent continuations and use the
non-overlapping grid
$\mathcal{T}_{\mathrm{eval}}^{\mathrm{syn}}
=\{H,\ H+T,\ H+2T,\ \dots,\ H+N-T\}$.
We then pool segments over continuation index $j=1,\dots,J$ and
evaluation times $t\in\mathcal{T}_{\mathrm{eval}}^{\mathrm{syn}}$,
flattening $(j,t)$ into a single empirical collection.

We assess the similarity between these empirical path distributions
using a large suite of metrics defined below.
Many of these metrics have been proposed and used in prior work
\citep{yoon_2019_times, ni_2022_sigwa, wiese_2021_multia, liao_2024_sigwa, cont_2025_tailg}.

\paragraph{Motivation of our training \& evaluation protocol}
Our single-path training and out-of-sample evaluation protocol is designed to mimic
how generative time series models are trained and deployed in financial applications,
where generative time series models are used as \emph{market simulators}
\citep{wiese_2019_deepa, buehler_2020_a, cohen_2021_black, lutkebohmert_2022_robus, wiese_2022_riskn, limmer_2024_robus, cont_2025_datad, jones_2025_ambig, he_2025_distr}.
These market simulators are often used to optimize and stress-test decision rules (e.g., via neural network policies)
for general portfolio management and hedging problems
\citep{buehler_2019_deep, mariani_2019_pagan, murray_2022_deep, krabichler_2023_a, englisch_2023_deep, mueller_2024_fast}.
Across all these applications, the core generative problem is the same:
given a single observed historical path, generate future trajectories conditional on the recent past
(\cref{subsec:problem_setting}).
To obtain sufficient out-of-sample samples for the distributional evaluation,
we train the generator once and keep it fixed throughout evaluation,
rolling it forward across evaluation times.
Such simulator reuse can also be useful in practice,
since it amortizes generator training cost
and because incremental new data often has limited effects on the optimal generator.
In practice, the generator may nonetheless be retrained
once enough new data is available or after detecting a distributional shift.
For simplicity and to control compute, we do not perform such generator refitting.

\paragraph{Broader impacts}\label{app:par:broader_impacts}
Improved financial time series generation from small datasets can support
risk analysis and stress testing for many financial applications
when historical data is limited.
The same capability also creates a risk of overreliance if
uncertainty about the generative model is ignored.

\subsubsection{Discriminative evaluation metrics}

\begin{table}[t]
    \centering
    \small
    \caption{Evaluation metrics. Discriminative metrics report
    $|\mathrm{accuracy}-0.5|$ for classifiers trained to distinguish real from
    generated paths. All metrics are defined so that lower values indicate better fit.}
    \label{tab:metrics}
    \vspace{0.5em}
    \begin{tabular}{@{}lp{0.78\linewidth}@{}}
        \toprule
        Metric & Description \\
        \midrule
        \multicolumn{2}{@{}l}{\emph{Discriminative metrics}} \\
        \textsc{SIG}
        & Linear classifier on truncated signature features. \\
        \textsc{MLP}
        & Linear classifier on random one-hidden-layer MLP features of flattened paths. \\
        \textsc{RNN}
        & GRU classifier applied to individual paths, following \citet{yoon_2019_times}. \\
        \textsc{SRNN}
        & Permutation-invariant GRU classifier applied to sets of 8 paths. \\
        \cmidrule(lr){1-2}
        \multicolumn{2}{@{}l}{\emph{Distributional metrics}} \\
        \textsc{ACF}
        & Average absolute discrepancy between channel-wise autocorrelation functions. \\
        \textsc{CCF}
        & Average absolute discrepancy between Pearson cross-correlation matrices. \\
        \textsc{CVM}
        & Average Cram\'er--von Mises distance between one-dimensional channel marginals. \\
        \textsc{ES}
        & Average relative discrepancy in per-channel expected shortfall at level $5\%$. \\
        \textsc{ED}
        & Energy distance between empirical distributions of consecutive observation
        pairs $(X_t, X_{t+1})\in\mathbb{R}^{2d}$. \\
        \textsc{PES}
        & Average relative discrepancy in expected shortfall of trading-strategy
        PnL distributions, following \citet{cont_2025_tailg}. \\
        \bottomrule
    \end{tabular}
\end{table}

We report four discriminative metrics:
\textsc{SIG},
\textsc{MLP},
\textsc{RNN},
and \textsc{SRNN}.
For each metric, we train a binary
classifier to distinguish real paths from fake paths.
We train on half of the validation set, evaluate accuracy on the other half,
and report the discriminative score $\lvert \mathrm{accuracy} - 0.5 \rvert$.

For the metrics \textsc{SIG} and \textsc{MLP},
we fit linear ridge classifiers on top of fixed feature maps.
\textsc{SIG} uses truncated signature features (degree $3$),
computed on augmented versions of the paths
(same augmentations as described in \cref{app:subsec:baseline_details}).
\textsc{MLP} uses random features obtained by passing the flattened path
$\mathrm{vec}(X)\in\mathbb{R}^{Td}$
through a randomly initialized one-hidden-layer MLP (hidden dimension~$512$, ReLU).
To reduce evaluation variance due to the MLP initialization,
we repeat sampling the MLP and fitting the linear classifier for 5 times and report
the averaged discriminative score.

Following \citet{yoon_2019_times},
we further report \textsc{RNN}, which is the discriminative score
obtained by a fully trained GRU-based classifier \citep{chung_2014_empira}.
Note that these three discriminative scores
(\textsc{SIG}, \textsc{MLP}, \textsc{RNN})
classify individual paths as real or generated.
To better capture distributional properties,
we also report \textsc{SRNN}, which classifies \emph{sets} of paths rather than individual paths.
Concretely, given a set of $B_S = 8$ paths $\{X^{i} \}_{i=1}^{B_S}$,
we compute a GRU embedding for each path and then apply a Deep Sets aggregator \citep{zaheer_2017_deep}:
\begin{equation*}
\ell =
\rho\Bigl(\frac{1}{B_S}\sum\nolimits_{i=1}^{B_S} \phi(\mathrm{GRU}(X^i))\Bigr),
\qquad
p(\text{real}\mid \{X^i\}_{i=1}^{B_S}) = \sigma(\ell).
\end{equation*}
Here $\phi$ and $\rho$ are MLPs,
$\sigma$ is the sigmoid function,
and the mean-pooling makes the score permutation-invariant over the set.
To reduce training noise,
we train two GRU-based classifiers
(\textsc{RNN} and \textsc{SRNN})
ten times and average the resulting scores.

\subsubsection{Distributional evaluation metrics}
Below, superscripts $r$ and $g$ denote quantities computed on the real and generated segment collections, respectively.
We report three metrics (\textsc{CVM}, \textsc{CCF}, \textsc{ES})
that ignore temporal ordering and for which we aggregate all observations
across segments and time.
The other three metrics (\textsc{ACF}, \textsc{ED}, \textsc{PES})
capture temporal dependencies.

\textbf{Cramér--von Mises distance (\textsc{CVM}).}
We measure similarity of the one-dimensional marginals using the average Cramér--von Mises distance:
\begin{equation*}
\text{\textsc{CVM}}
=
\frac{1}{d} \sum_{j=1}^{d}
\int_{-\infty}^{\infty}
\left(
\hat{F}^{(r, j)}(u) - \hat{F}^{(g, j)}(u)
\right)^2
\, \mathrm{d}\hat{H}^{(j)}(u),
\end{equation*}
where $\hat{F}^{(r, j)}$ and $\hat{F}^{(g, j)}$ are the ECDFs of channel $j$ computed from the aggregated real and generated observations, and $\hat{H}^{(j)}$ is the ECDF of the pooled real+generated sample in channel $j$.

\textbf{Cross-correlation difference (\textsc{CCF}).}
We measure cross-sectional dependence via the average absolute difference between Pearson cross-correlation matrices:
\begin{equation*}
\text{\textsc{CCF}}
=
\frac{1}{d(d-1)}\sum_{i\neq j}
\left|
\chi^{(r)}_{ij}-\chi^{(g)}_{ij}
\right|,
\end{equation*}
where $\chi^{(r)}, \chi^{(g)} \in \mathbb{R}^{d \times d}$ are the Pearson correlation matrices
computed from the aggregated real and generated observations, respectively.

\textbf{Relative expected shortfall difference (\textsc{ES}).}
We compare tail risk using the average relative absolute error in Expected Shortfall (ES) at level $\alpha = 0.05$:
\begin{equation*}
\text{\textsc{ES}}
= \frac{1}{d}\sum_{j=1}^{d}
\tfrac{\left|\mathrm{ES}^{(r)}_{\alpha,j}-\mathrm{ES}^{(g)}_{\alpha,j}\right|}
{\left|\mathrm{ES}^{(r)}_{\alpha,j}\right|}.
\end{equation*}
Here $\mathrm{ES}^{(\cdot)}_{\alpha,j}$ denotes the empirical expected shortfall of channel $j$ at tail level $\alpha$, computed as the mean of observations below the empirical $\alpha$-quantile (after aggregating over segments and time).

\textbf{Autocorrelation difference (\textsc{ACF}).}
We measure temporal dependence via the average absolute difference between channel-wise autocorrelation functions:
\begin{equation*}
\text{\textsc{ACF}}
=
\frac{1}{d\,\ell_{\max}}
\sum_{j=1}^{d}
\sum_{k=1}^{\ell_{\max}}
\left|
\rho^{(r)}_{j}(k)-\rho^{(g)}_{j}(k)
\right|,
\qquad
\ell_{\max}=\big\lfloor T/3 \big\rfloor,
\end{equation*}
where $\rho^{(r)}_{j}(k)$ and $\rho^{(g)}_{j}(k)$ denote the empirical autocorrelation at lag $k$ for channel $j$, computed from the real and generated segment collections, respectively.

\textbf{Energy distance (\textsc{ED}).}
We compute the Energy Distance between the empirical distributions of consecutive time-step pairs.
Let $(x,y) \in \mathbb{R}^{2d}$ denote two \emph{consecutive} observations from the same segment,
i.e., $y$ is the next time step after $x$:
\begin{equation*}
Z := (x,y) \in \mathbb{R}^{2d},
\qquad
\text{\textsc{ED}}
=
\sqrt{
2 \, \mathbb{E}\| Z^{(r)} - Z^{(g)} \|_2
- \mathbb{E}\| Z^{(r)} - Z'^{(r)} \|_2
- \mathbb{E}\| Z^{(g)} - Z'^{(g)} \|_2
},
\end{equation*}
where the expectations are taken with respect to the empirical distributions
of real and generated consecutive pairs (aggregated over segments and valid time steps).

\textbf{PnL expected shortfall difference (\textsc{PES}).}
Following \citet{cont_2025_tailg}, we assess tail fit
by comparing the tails of profit-and-loss (PnL) distributions
induced by simple trading strategies applied to real versus generated segments.
For each strategy $j\in\{1,\dots,N_{\text{strategies}}\}$,
we compute a PnL time series on the real and generated data,
denoted $\mathrm{PnL}_j^{(r)}$ and $\mathrm{PnL}_j^{(g)}$,
and compare their expected shortfall at tail level $\alpha = 0.05$.
We define
\begin{equation*}
\text{\textsc{PES}}
=
\frac{1}{N_{\text{strategies}}}
\sum_{j=1}^{N_{\text{strategies}}}
\tfrac{\left|\mathrm{ES}_\alpha\!\left(\mathrm{PnL}_j^{(r)}\right)
-\mathrm{ES}_\alpha\!\left(\mathrm{PnL}_j^{(g)}\right)\right|}
{\left|\mathrm{ES}_\alpha\!\left(\mathrm{PnL}_j^{(r)}\right)\right|}.
\end{equation*}
We use the following strategies:
\begin{itemize}
\item \emph{Buy-and-hold (single channel):} $d$ strategies, each taking a unit position $+1$ in a single channel and $0$ in all others.
\item \emph{Buy-and-hold ($1/d$):} one strategy taking an equal-weight position $1/d$ in each channel.
\item \emph{Momentum:} $d$ strategies, where the position in channel $j$ at time $t$ is $\mathrm{sign}(\bar{X}^{(j)}_t)$, with $\bar{X}^{(j)}_t$ the 10-step moving average of channel $j$ up to time $t$.
\end{itemize}

\subsection{Additional training details}\label{app:subsec:training_details}

\subsubsection{Data and feature scaling}\label{app:subsubsec:feature_scaling}
\textbf{Data scaling.} On each dataset, we compute the per-channel means and standard deviations
on the raw training path and use those vectors
to standardize the training and validation data.

\textbf{Feature scaling.}
We optimize the objective in \cref{eq:random_feature_matching_objective}
using minibatches of size $B$ for the inner expectations
and one draw of the feature-map parameters $\psi$ at a time.
At each step, we sample indices $t_1,\dots,t_B$ uniformly from
the valid training indices and generate
$\hat x_{t_i}^+ \sim p_\theta(\cdot\mid x_{t_i}^-)$.
We then compute
\begin{equation*}
F_i = f_\psi(x_{t_i}^- \oplus x_{t_i}^+),
\qquad
\hat F_i = f_\psi(x_{t_i}^- \oplus \hat x_{t_i}^+).
\end{equation*}
The corresponding minibatch version of \cref{eq:random_feature_matching_objective}
is therefore
\begin{equation*}
\mathcal L_B(\theta;\psi)
=
\biggl\|
\frac{1}{B}\sum_{i=1}^B F_i
-
\frac{1}{B}\sum_{i=1}^B \hat F_i
\biggr\|_2^2 .
\end{equation*}
For all feature maps (\texttt{SOCK}, \texttt{rSig}, and \texttt{Sig}),
we also considered the scaled loss
\begin{equation*}
\mathcal L_B^{\mathrm{scaled}}(\theta;\psi)
=
\biggl\|
\boldsymbol{\sigma}_\psi^{-1}
\odot
\biggl(
\frac{1}{B}\sum_{i=1}^B F_i
-
\frac{1}{B}\sum_{i=1}^B \hat F_i
\biggr)
\biggr\|_2^2 ,
\end{equation*}
where $\boldsymbol{\sigma}_{\psi,j}$ is the empirical standard deviation
of feature coordinate $j$ over all real training segments for the current
feature-map draw, and $\odot$ denotes componentwise multiplication.
This scaling is related to feature matching methods that use higher moments \citep{mroueh_2017_mcgan}.
We found in preliminary experiments that this scaling improved generator
training with SOCK features.
However, this scaling does not generally improve generator training with
\texttt{Sig} or \texttt{rSig} features
(\cref{tab:appendix_ablation_feature_scaling}).
In all experiments, we therefore apply this scaling to \texttt{SOCK} features,
but do not apply it to \texttt{Sig} or \texttt{rSig} features,
which is consistent with prior works
\citet{liao_2024_sigwa,2024NiklasGononUniversalrandomisedsignaturesgenerative}.

\begin{table}[t]
\caption{
Feature-scaling ablation for \texttt{Sig} and \texttt{rSig} on the
\texttt{TGH} dataset; metrics are averaged over five random seeds.}
\label{tab:appendix_ablation_feature_scaling}
\begin{center}
\begingroup
\setlength{\tabcolsep}{3.5pt}
\scshape
\begin{small}
\begin{tabular}{lcrrrrrr}
\toprule
Method & Scaling & \makebox[2.0em]{RNN} & \makebox[2.0em]{ACF} &
\makebox[2.0em]{CCF} & \makebox[2.0em]{CVM} &
\makebox[2.0em]{ED} & \makebox[2.0em]{ES} \\
\midrule
\texttt{Sig}  & $\times$     & \textbf{0.319} & \textbf{0.079} & 0.032 & \textbf{0.0008} & \textbf{0.079} & \textbf{0.109} \\
\texttt{Sig}  & \checkmark   & 0.368 & 0.084 & \textbf{0.025} & 0.0010 & 0.101 & 0.110 \\
\cmidrule(lr){1-8}
\texttt{rSig} & $\times$     & 0.296 & 0.089 & 0.062 & \textbf{0.0003} & \textbf{0.068} & \textbf{0.044} \\
\texttt{rSig} & \checkmark   & \textbf{0.288} & \textbf{0.085} & \textbf{0.049} & 0.0005 & 0.084 & 0.047 \\
\bottomrule
\end{tabular}
\end{small}
\endgroup
\end{center}
\end{table}

\subsubsection{Generator architecture}\label{subsubsec:gen_arch}
We use a simple conditional generator based on a single-layer
GRU that decodes an i.i.d.\ noise sequence into an output path.
We choose this architecture for simplicity and fast sampling.
Improving the generator architecture is complementary
to our approach and may yield further gains.

\begin{tcolorbox}[codebox]
\begin{minted}[breaklines,autogobble,fontsize=\scriptsize]{python}
class Generator(nn.Module):
    def __init__(self, d: int, hidden_dim: int = 128, q: int = 5) -> None:
        super().__init__()
        self.noise_dim = self.initial_noise_dim = d  # set noise dimension to data dimension

        # maps (flattened context c, initial noise) -> initial hidden state h0
        self.initial_state_generator = nn.Sequential(
            nn.Linear(d * q + self.initial_noise_dim, hidden_dim),
            nn.SiLU(),
            nn.Linear(hidden_dim, hidden_dim),
        )

        self.proj_in = nn.Linear(self.noise_dim, 2 * hidden_dim)
        self.rnn = nn.GRU(hidden_dim, hidden_dim, num_layers=1, batch_first=True)

        # residual noise injection: add a gated noise stream after the GRU
        self.alpha = nn.Parameter(torch.tensor(0.1))
        self.gate = nn.Sequential(
            nn.LayerNorm(hidden_dim), nn.Linear(hidden_dim, hidden_dim), nn.Sigmoid()
        )

        self.proj_out = nn.Linear(hidden_dim, d)

    def reset_parameters(self) -> None: ...  # stabilizing initialization (e.g., small output/MLP scales)

    def forward(self, c: Tensor, n_steps: int = 64) -> Tensor:  # c.shape = (B, q, d)
        # sample initial hidden state h0 conditionally on context c
        initial_noise = torch.randn((c.size(0), self.initial_noise_dim), device=c.device)
        h0_in = torch.cat((c.flatten(start_dim=1), initial_noise), dim=-1)
        h0 = self.initial_state_generator(h0_in)

        # decode per-step noise with: linear proj. -> SiLU -> GRU
        z = torch.randn((c.size(0), n_steps, self.noise_dim), device=c.device)
        z, z_skip = self.proj_in(z).chunk(2, dim=-1)
        h, _ = self.rnn(F.silu(z), h0.unsqueeze(0))

        h = h + self.alpha * self.gate(h) * z_skip  # optional output noise injection
        return self.proj_out(h)
\end{minted}
\end{tcolorbox}

\paragraph{Do generators learn to condition on the past?}
To evaluate this architecture,
we test whether generators trained with \texttt{SOCK} features use their
conditioning context by comparing correct and randomly shuffled contexts.
Let $\{x_t^-\}$ be the evaluation contexts and let $\pi$ be a random
permutation of their indices. Normally, we sample
$\hat{x}_t^+ \sim p_\theta(\cdot \mid x_t^-)$ and evaluate the joined path
$x_t^- \oplus \hat{x}_t^+$. For the shuffled evaluation, we instead sample
$\widetilde{x}_t^+ \sim p_\theta(\cdot \mid x_{\pi(t)}^-)$ and evaluate
$x_t^- \oplus \widetilde{x}_t^+$. This gives two collections of paths:
$\mathcal{X}_{\mathrm{correct}} = \{x_t^- \oplus \hat{x}_t^+\}_t$, which
pairs each context with a continuation generated from that context, and
$\mathcal{X}_{\mathrm{shuffled}} = \{x_t^- \oplus
\widetilde{x}_t^+\}_t$, which pairs each context with a continuation generated
from the permuted context $x_{\pi(t)}^-$.
We compute the ACF metric on each collection as it is
sensitive to whether the generated continuation is temporally consistent
with the real prefix.
We perform this evaluation for \texttt{SOCK} generators trained on the \texttt{TGH} and
\texttt{SV} datasets (five seeds each). For each generator, we average results
over five draws of the permutation $\pi$.
Supplying the wrong contexts to the generators results in an increase
of the ACF metric by $18.2\%$ on \texttt{TGH} and $99.1\%$ on
\texttt{SV} (\cref{tab:appendix_context_shuffling}).
This shows that the trained generators have learned to meaningfully
use the conditioning context to produce future continuations.

\begin{center}
\captionof{table}{
Context-shuffling ablation for generators trained with \texttt{SOCK} features.
Results are averaged over five training seeds and five context-permutation and
noise realizations per seed.}
\label{tab:appendix_context_shuffling}
\begingroup
\setlength{\tabcolsep}{7pt}
\scshape
\begin{small}
\begin{tabular}{lrrr}
\toprule
Dataset & ACF on $\mathcal{X}_{\mathrm{correct}}$
& ACF on $\mathcal{X}_{\mathrm{shuffled}}$ & Relative increase \\
\midrule
\texttt{TGH} & \textbf{0.024578} & 0.029062 & $+18.2\%$ \\
\texttt{SV}  & \textbf{0.023370} & 0.046539 & $+99.1\%$ \\
\bottomrule
\end{tabular}
\end{small}
\endgroup
\end{center}

\subsubsection{Optimization}
For all the feature matching methods
(\texttt{Sig},
\texttt{rSig},
\texttt{cSig},
\texttt{SOCK}),
we use the same generator architecture
and training protocol across all datasets.
We tuned these choices on the synthetic datasets
to avoid overfitting to the small real datasets.
We train the generator for 100,000 optimization steps
using the AdamW optimizer (weight decay of $0.01$)
and a linear learning rate schedule
(linear warm up for first 5\% of steps to $3 \times 10^{-4}$
and linear decay for last 70\% of steps).

\paragraph{Compute.}
All reported generative experiments fit on a single NVIDIA A40 GPU with 48GB memory.
For one dataset and seed,
training a GRU generator with feature matching for 100k steps takes about 1 hour;
sampling 65k paths takes less than 1 minute.
Diffusion-TS trains faster, about 10 minutes,
but sampling 65k paths takes more than 30 minutes.
\texttt{SOCK} feature extraction is not the training bottleneck;
the dominant cost is the backward pass through the GRU generator.

\subsection{Additional details for the baselines}\label{app:subsec:baseline_details}

\subsubsection{Truncated signature (\texttt{Sig})}

As discussed in \cref{subsec:gen_models},
we apply three path transforms prior to computing
signature terms for all signature-based methods:
Lead-Lag,
Time,
and I-visibility.
To compute the truncated signature,
we use Signatory \citet{kidger2021signatory}.
We truncate the signature of the augmented path at degree $3$,
following \citet{liao_2024_sigwa}.

\subsubsection{Randomized signature (\texttt{rSig})}

Due to the exponential $\mathcal{O}(d^m)$ time and space complexity of computing $m$-level truncated signatures for $d$-dimensional paths or time series, the randomized signature \cite{2021ExpressivePowerOfRandomizedSignatures, 2023NeuralSignatureKernels} is often preferred, defined below.

\begin{definition}\label{defRandomizedSignature}
    Let $M\geq 1$ be an integer. Fix an initial condition $z_0 \in \mathbb{R}^M$, random matrices $A_1, \cdots, A_d \in \mathbb{R}^{M\times M}$, random biases $b_1, \cdots, b_d \in \mathbb{R}^{M}$ and an activation function $\sigma$. The randomized signature $Z$ of a path $t \mapsto x_t$ is defined as the solution of the controlled differential equation (CDE)
    \begin{align}\label{eqRandomizedSignature}
        dZ_t = \sum_{i=1}^d \sigma(A_i Z_t + b_i)dx_t^{(i)}, \qquad Z_0 = z_0,
    \end{align}
    where $x^{(i)}$ denotes the $i$'th component of $x$.
\end{definition}

The randomized signature was first constructed by \citet{2021ExpressivePowerOfRandomizedSignatures} as a random projection of the signature, with an argument based on a non-trivial application of the Johnson-Lindenstrauss lemma. Randomized signatures have previously been used for various time series applications, for instance anomaly detection \citep{2022CrpytoMarketAnomalyDetection, 2025InfiniteDimensionalMahalanobisDistance}, graph conversion \citep{schäfl2023gsignatures}, optimal portfolio selection \citep{2023TeichmannOptimalPortfolioSelection,2025cuchieroSignaturePortfolioTheory}, generative
modeling \citep{2024NiklasGononUniversalrandomisedsignaturesgenerative}, and for learning rough dynamical systems \citep{2023OnTheEffectivenessOfRandomizedSignaturesAsReservoirForLearningRoughDynamics}. The CDE \eqref{eqRandomizedSignature} has since been studied from the perspective of randomly initialized ResNets \citep{2023NeuralSignatureKernels, 2024NikolaTheoreticalDoundationsDeepSelectiveStateSpaceModels}, and path developments on compact Lie groups \citep{2023PCFGAN, 2024PathDevelopmentNetworkWithFiniteDimensionalLieGroup, 2024WillTurnerThomasCassFreeProbabilityPathDevelopments}. We use the version of \citet{2023NeuralSignatureKernels}
with vector fields $A_i\sigma(Z_t)+b_i$ and $\sigma=\tanh$,
implemented below.
In preliminary experiments, we observed similar performance
to an Euler discretization of \eqref{eqRandomizedSignature}.
We expect differences in generator performance between the
two versions to be small, particularly when the random
parameters are resampled during generator training.

\begin{tcolorbox}[codebox]
\begin{minted}[breaklines,autogobble,fontsize=\scriptsize]{python}
def randomized_sig_tanh(X: Tensor, A: Tensor, b: Tensor, Y_0: Tensor) -> Tensor:
    """
    Randomized signature of a (batched) time series X, with tanh activation function.

    Args:
        X (Tensor): Input tensor of shape (N, T, D).
        A (Tensor): Tensor of shape (M, M, D). Random matrix.
        b (Tensor): Tensor of shape (M, D). Random bias.
        Y_0 (Tensor): Initial hidden state.
            Tensor of shape (M).
    """
    N, T, D = X.shape
    diff = X.diff(dim=1) # shape (N, T-1, D)
    Y_0 = torch.tile(Y_0, (N, 1)) # shape (N, M)
    Z = torch.tensordot(tanh(Y_0), A, dims=1) + b[None] # shape (N, M, D)
    Y = Y_0 + (Z * diff[:, 0:1, :]).sum(dim=-1) # shape (N, M)
    for t in range(1, T-1):
        Z = torch.tensordot(tanh(Y), A, dims=1) + b[None]
        Y = Y + (Z * diff[:, t:t+1, :]).sum(dim=-1)
    return Y
\end{minted}
\end{tcolorbox}

\subsubsection{Conditional Sig-Wasserstein GAN (\texttt{cSig})}
\citet{liao_2024_sigwa}
train conditional generators by matching signature features of future
segments.
For each training time $t$, we distinguish three segments:
\begin{equation*}
    x_t^- := X_{t-q+1:t},
    \qquad
    r_t^- := X_{t-T_-+1:t},
    \qquad
    x_t^+ := X_{t+1:t+T_+}.
\end{equation*}
The generator receives the $q$ observations in $x_t^-$ and samples
$\hat{x}_t^+ \sim p_\theta(\cdot \mid x_t^-)$.
The longer segment $r_t^-$ is used only to construct the target feature mean
for the real future segment $x_t^+$.

Let $f$ denote the truncated signature of an augmented input path.
If the target feature mean of the real future were available, the \texttt{cSig} loss
would compare it to the generator's conditional feature mean:
\begin{equation}
\mathbb{E}_{t}\Bigl[
\bigl\|
\mathbb{E}_{x_t^+ \sim p(\cdot \mid r_t^-)}
    \bigl[f(x_t^+)\bigr]
-
\mathbb{E}_{\hat{x}_t^+ \sim p_\theta(\cdot \mid x_t^-)}
    \bigl[f(\hat{x}_t^+)\bigr]
\bigr\|_2^2
\Bigr] .
\end{equation}
The second expectation can be estimated by Monte Carlo, because we can sample
many generated futures from $p_\theta(\cdot \mid x_t^-)$.
The first expectation is not directly observable from a single historical path:
for each $r_t^-$ we observe only one realized future $x_t^+$.

\citet{liao_2024_sigwa} therefore estimate the first expectation before
training the generator.
They fit a linear map from signature features of the longer past segment
to signature features of the future segment,
\begin{equation*}
\mathbb{E}_{x_t^+ \sim p(\cdot \mid r_t^-)}
    \bigl[f(x_t^+)\bigr]
\approx b + W f(r_t^-) .
\end{equation*}
The coefficients $(b,W)$ are fitted on the training path and then held fixed.
The generator is then trained with the practical objective
\begin{equation*}
\mathbb{E}_{t}\Bigl[
\bigl\|
b + W f(r_t^-)
-
\mathbb{E}_{\hat{x}_t^+ \sim p_\theta(\cdot \mid x_t^-)}
    \bigl[f(\hat{x}_t^+)\bigr]
\bigr\|_2^2
\Bigr] .
\end{equation*}
Thus the target in the \texttt{cSig} loss is predicted from $r_t^-$, while the
generator itself is still conditioned only on $x_t^-$.
This objective differs from ours in \cref{eq:random_feature_matching_objective}:
\texttt{cSig} matches conditional feature means of the future segment
$x_t^+$, whereas our objective matches features of the joined segment
$x_t^- \oplus x_t^+$.

\citet{liao_2024_sigwa} choose leading and generated segments of equal length
($T_- = T_+$) in their experiments.
We make the same choice for our \texttt{cSig} generators:
$T_- = T_+ = T = 64$.
Thus the regression map is fitted from $T$ past observations,
while the generator is still conditioned on the $q=5$ observations in
$x_t^-$.

All runs with \texttt{cSig} therefore use $T_- = T_+ = 64$.
We implement the nested sampling scheme at batch size
$B = 1000$:
for each of the $B^{1/3} = 10$ outer samples we use $B^{2/3} = 100$ inner samples
to approximate the inner expectations \citep{gordy_2010_nestea}.

\subsubsection{Diffusion-TS (\texttt{DTS})}\label{subsubsec:dts}
We train a separate Diffusion-TS model for each dataset and each seed.
We adapt the authors' implementation \citep{yuan_2024_diffu} to run on our datasets.\footnote{
See \url{https://github.com/Y-debug-sys/Diffusion-TS}.
}
Note that by default, Diffusion-TS uses the same stride-$1$ sampling scheme
for training to extract smaller path segments from a single historical time series as we do.
We followed \citet{yuan_2024_diffu}'s preprocessing choices for our datasets,
which require scaling the training data to the unit interval~$[0, 1]$.
During evaluation, Diffusion-TS again produces only samples in this range,
which may limit performance of Diffusion-TS on our out-of-sample evaluation.

To assess this support restriction, we compare the default sampling procedure
with sampling that disables clipping of the predicted clean sample to the model
range $[-1,1]$ (by setting \texttt{clip\_denoised=False}).
We run this ablation on \texttt{TGH} using the same trained models for both
sampling procedures.

\begin{center}
\captionof{table}{
Effect of sample clipping for Diffusion-TS on the \texttt{TGH} dataset;
metrics are averaged over two seeds.}
\label{tab:appendix_dts_clipping}
\begingroup
\setlength{\tabcolsep}{4pt}
\scshape
\begin{small}
\begin{tabular}{lrrrrrrr}
\toprule
Method & RNN & ACF & CCF & CVM & ED & ES & PES \\
\midrule
\texttt{DTS} (clipping)    & \textbf{0.382} & 0.051 & 0.077 & \textbf{0.0026} & \textbf{0.256} & \textbf{0.304} & \textbf{0.162} \\
\texttt{DTS} (no clipping) & 0.454 & \textbf{0.037} & \textbf{0.060} & 0.0072 & 0.386 & 0.409 & 1.055 \\
\bottomrule
\end{tabular}
\end{small}
\endgroup
\end{center}

Disabling clipping improves ACF and CCF but worsens the other reported metrics
(\cref{tab:appendix_dts_clipping}).
In particular, both tail metrics, ES and PES, deteriorate when clipping is disabled.
A possible explanation for this is that the generator produces extreme outliers
when clipping is disabled.
Clipping then suppresses these samples,
which improves the tail metrics despite biasing the generated tails.

\clearpage
\section{Additional details and ablations on discriminative experiments} \label{app:sec:discrimative_eval}

\subsection{Hypothesis testing}\label{app:subsec:hypotesting}

We illustrate the expressive power of SOCK features
on a two-sample hypothesis test for stochastic processes.
Specifically, we use the maximum mean discrepancy (MMD) \citep{2012MMDAKernelTwoSampleTest}
based on either explicit feature vectors
(as in \cref{eq:feature_matching_loss}) or time series kernels,
within a permutation test as in
\citet{2022SignatureMomentsToCharacterizeLawsOfStochasticProcesses,2021HigherOrderKernelMeanEmbeddings,2023PCFGAN,2024RestrictedPathCharacteristicFunctionDeterminesTheLawOfStochasticProcesses,2024HRPCFGAN}.
We test the null hypothesis $H_0: \mu = \nu$ against the alternative $H_1: \mu \neq \nu$, where $\mu$ and $\nu$ denote laws of stochastic processes. 

\paragraph{Test Procedure}
Let $\mathcal{X}=(X^{(1)}, \dots, X^{(N)}) \sim \mu$
and $\mathcal{Y}=(Y^{(1)}, \dots, Y^{(M)}) \sim \nu$
be independent samples of sizes $N$ and $M$, respectively, with i.i.d.\ elements.
The permutation test proceeds as follows:  
\begin{enumerate}
    \item
    Let $\mathcal{Z} = (X^{(1)}, \dots, X^{(N)}, Y^{(1)}, \dots, Y^{(M)})$.
    Randomly partition $\mathcal{Z}$ into sets
    $\widetilde{\mathcal{X}}$ and $\widetilde{\mathcal{Y}}$ of sizes $N$ and $M$ respectively.
    Compute and record $\text{MMD}(\widetilde{\mathcal{X}}, \widetilde{\mathcal{Y}})$. Repeat 100 times.
    \item Compute the empirical $(1-\alpha)$ quantile of the permuted MMD values, which under $H_0$ has the same distribution as the unpermuted statistic $\text{MMD}(\mathcal{X}, \mathcal{Y})$.
    \item Reject $H_0$ if $\text{MMD}(\mathcal{X},\mathcal{Y})$ exceeds this threshold.  
\end{enumerate}

This procedure is repeated for 300 trials to estimate the test power. We consider three different stochastic processes that capture different time series characteristics: 1.) \texttt{FBM}: We vary the Hurst exponent $H \in (0,1)$ under $H_1$, and set $H=0.5$ under $H_0$. 2.) \texttt{TGH}: We vary the tail heaviness parameter $h$ under $H_1$, and set $h=0$ under $H_0$. 3.) \texttt{SV}: We vary the correlation $\rho \in (-1, 1)$ under $H_1$, and set $\rho=0$ under $H_0$.

\paragraph{Hyperparameter Tuning}
For each stochastic process, we independently tune the hyperparameters of each MMD model to maximize the test power, using a hold-out data set generated using a separate random seed. Hydra uses its default hyperparameters and requires no tuning as per the original paper \cite{2023Hydra}. We use a grid search to tune the $\sigma$ parameter for all RBF-based models, including GAK, and to determine the optimal truncation depth for all truncated signature MMDs. For the randomized signature, we tune the path dimension and the variance of the random maps. For \texttt{SOCK} and Hydra we set the number of competing kernels to $k=3$, and for \texttt{SOCK} we only tune the softmax temperature, but we note that this only had a small increase in performance. We stress that \texttt{SOCK}, like its predecessor Hydra, is robust to changes in hyperparameters and we recommend using the defaults for new problems. 
Additionally, for all signature-based models we preprocess all paths with the standard path augmentations from the signature literature, namely Lead-lag, Add-time, and I-visibility transformations \cite{2025SignatureMethodsInMachineLearning, 2020SignatureFeaturesWithTheVisibilityTransformation}. We note however that this has little effect on \texttt{SOCK}, but is required to make signatures competitive. The experiment results are displayed in  \cref{fig:hypothesis-testing}.

\subsection{Time series classification}\label{app:subsec:tsc}

In this section we provide additional experimental details and ablation studies for the time series classification evaluation on the classical 112 UCR datasets \citep{2024BakeOff}.
The UCR Time Series Classification Archive is distributed under CC-BY 4.0.\footnote{
\url{https://zenodo.org/records/11198697}.
}

\paragraph{Experimental Setup} For all classification experiments, we follow the standard evaluation protocol established by the Rocket and Hydra literature \citep{2020Rocket, 2021MiniRocket, 2022MultiRocket, 2023Hydra}. We generate random convolutional features using either \texttt{SOCK}'s softmax-deviation pooling or the pooling mechanisms from Rocket/Hydra, and then fit a RidgeCV classifier (see e.g. \texttt{scikit-learn} \citep{2011ScikitLearn}). We use the standard training and test splits provided by the UCR archive \citep{2024BakeOff}, reporting results averaged across 30 seeded resamples. We perform two types of experiments:

\begin{enumerate}
    \item \textbf{Controlled ablations and pooling comparisons:} We keep the preprocessing backbone fixed for all models and parameter choices, varying only one parameter at a time. This includes ablating the specific pooling operations used by SOCK, Rocket, and Hydra, measuring the effect of random bias, and analyzing SOCK's sensitivity to softmax temperature $\tau$ and the number of competing kernels $k$. For these experiments, we use only the \texttt{diff} input path augmentation to ensure consistency with the standard time series classification literature.

    \item \textbf{State-of-the-art comparison:} We compare our best \texttt{SOCK} classifier against the top-performing time series classifiers in each model category as identified by \citet{2024BakeOff}. We separate this from the controlled ablation because the SOTA baselines (like MultiRocket and Hydra) use slightly different internal preprocessing steps than our fixed backbone.
\end{enumerate}

\paragraph{Pooling comparison}
In the right panel of \cref{fig:classification}
we compare different choices for the temporal pooling from the Hydra and Rocket literature.
To ensure a fair comparison between random convolutional models, we use the exact same pipeline for each, changing only the pooling operation.

Hydra \citep{2023Hydra} forms two features from each group of competing kernels:
it counts how often each kernel wins the local competition
and accumulates the corresponding winning response values.
In \citet{2023Hydra}, these poolings are called ``hard counting'' and ``soft counting''.
We refer to these two poolings as \texttt{h-count} and \texttt{h-value} respectively.
And we refer to differentiable analogues of these two features,
introduced and defined by us in Appendix~\ref{app:subsec:sock_pooling},
as soft-count (\texttt{s-count}) and soft-value (\texttt{s-value}) respectively.
Here, ``soft'' stands for the $\mathrm{softmax}$ operation used as part of these poolings;
it is not related to the ``soft'' terminology used by \citep{2023Hydra}.
We further compare to Rocket's \texttt{max} and \texttt{ppv} poolings,
and MultiRocket's \texttt{ppv}, \texttt{mpv}, \texttt{mipv}, and \texttt{lspv} poolings.
With some minor abuse of notation, we refer to both the \texttt{SOCK} feature map and the resulting classifier as "\texttt{SOCK}".

\textbf{Single vs. multiple poolings.}

\begin{wrapfigure}{r}{0.55\linewidth}
    \centering
    \includegraphics[width=\linewidth]{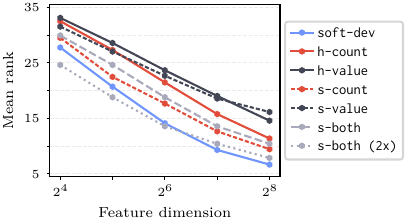}
    \caption{Mean rank on UCR classification tasks for different poolings:
    \texttt{SOCK}'s soft-deviation pooling (\texttt{soft-dev}),
    Hydra's count and value features (\texttt{h-count}, \texttt{h-value}),
    the differentiable analogues of Hydra's features
    (\texttt{s-count}, \texttt{s-value}),
    and their concatenation
    (\texttt{s-both}, which concatenates \texttt{s-count} and \texttt{s-value}).
    The \texttt{s-both} curve is compared at the same output feature dimension as the single-pooling curves, so it uses half as many convolution groups.
    The \texttt{s-both (2x)} curve uses the same number of convolution groups as the single-pooling curves and therefore has twice the output feature dimension.}
    \label{fig:single_vs_multiple_poolings}
\end{wrapfigure}

We use a single temporal pooling for summarizing
the soft kernel competitions in \texttt{SOCK},
while most other random convolutional feature maps use multiple poolings.
In particular, Hydra concatenates features from two poolings.
Concatenating multiple poolings could further improve
the \texttt{SOCK} feature map, and therefore improve classifiers or
generative models trained by feature matching.
However, \texttt{soft-dev} already performs strongly as a single pooling
(\cref{fig:classification}),
and concatenating multiple soft poolings does not always improve performance.

To see this, consider the result shown in
\cref{fig:single_vs_multiple_poolings}, which compares single soft and Hydra-style
poolings with concatenated poolings.
When matched by the effective number of output features,
combining soft-count and soft-value
(\texttt{s-both})
performs worse than using more soft-count features alone
(\texttt{s-count}).
This comparison is not cost-matched:
\texttt{s-count} uses more convolution groups to reach the same output dimension.
We therefore also compare to \texttt{s-both (2x)}, which uses both poolings
with the same number of convolution groups as the single-pooling curves
and therefore doubles the output dimension.
At small feature dimensions, this larger concatenated feature map outperforms
\texttt{soft-dev}.
At larger feature dimensions, the advantage disappears, and
\texttt{soft-dev} attains the better mean rank.

This comparison is meant to show that adding more poolings does not by itself
imply better performance.
It does not rule out gains from ensembling several poolings or several independent
feature maps, either for classification or for generative modeling.
It does show that \texttt{soft-dev} is competitive on its own, which motivates
our focus on the simpler single-pooling feature map.

\begin{figure*}
    \centering
    \includegraphics[width=\linewidth]{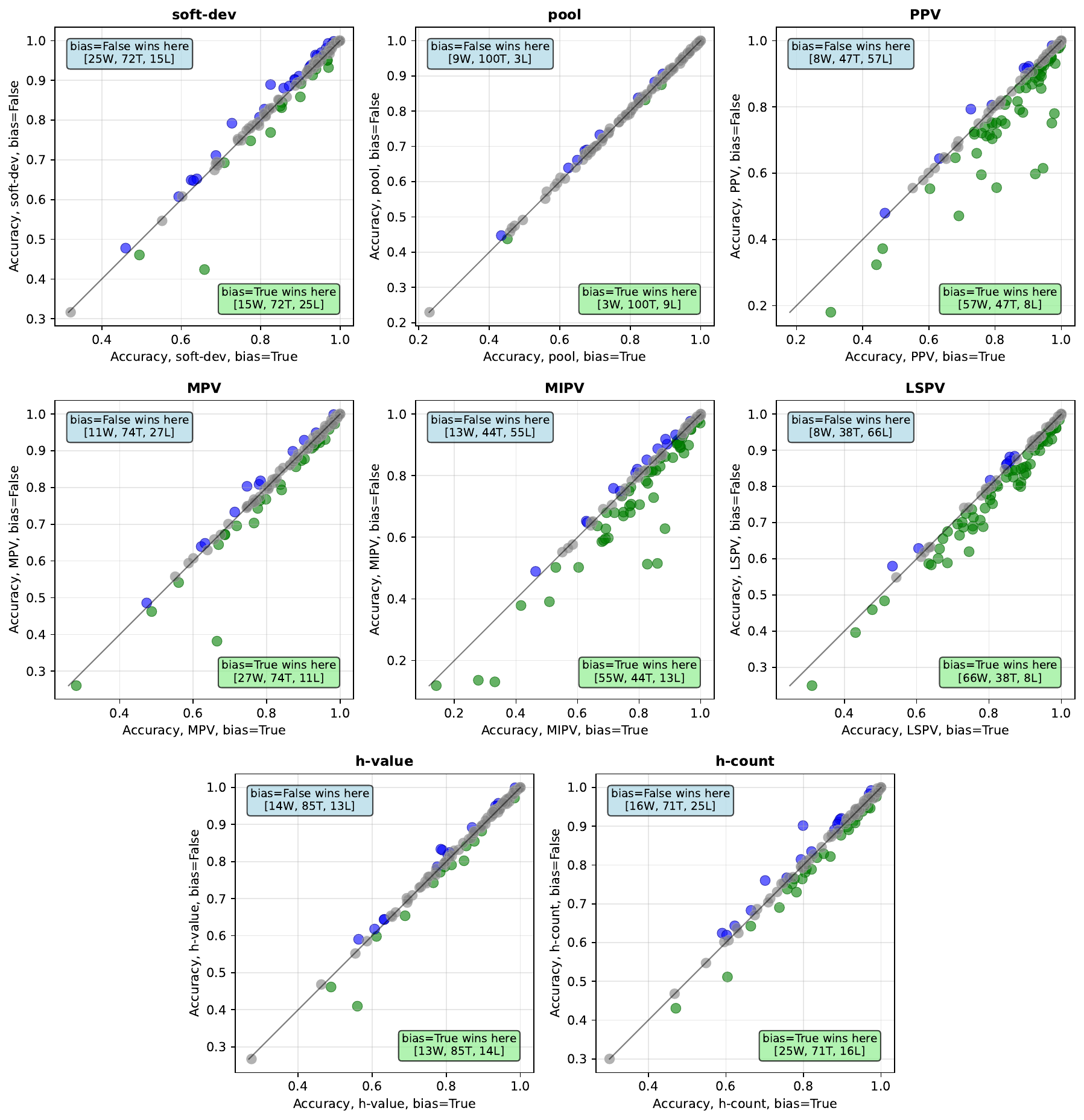}
    \caption{The effects of random bias for each pooling operation used by the random convolutional models SOCK, Rocket, MultiRocket, and Hydra.
    Results are reported as mean accuracies across 30 resamples on the 112 datasets of the UCR time series classification repository. Each point in the pairwise scatter plot represents one dataset. Accuracies within 0.01 are reported as ties in gray.}
    \label{fig:bias_vs_no_bias_comparison_allstats}
\end{figure*}

\textbf{Random bias.}
\Cref{fig:bias_vs_no_bias_comparison_allstats} presents pairwise scatter plots comparing each random pooling mechanism with and without the addition of a random bias term to the convolution output. We observe that \texttt{SOCK} and Hydra are, on average, insensitive to the addition of random bias. The \texttt{max} pooling is also naturally unaffected. In contrast, \texttt{ppv}, \texttt{mpv}, \texttt{mipv}, and \texttt{lspv} rely heavily on the random bias term for performance. Consequently, we include random bias for these pooling methods in our ranking comparisons. As shown in \cref{fig:classification}, \texttt{SOCK} outperforms all other individual pooling operations when other parameters are held fixed. Notably, \texttt{SOCK} with only 512 random features matches the performance of the otherwise best pooling \texttt{ppv} using 4096 features.

\begin{figure}[h]
    \centering
    \includegraphics[width=0.49\linewidth]{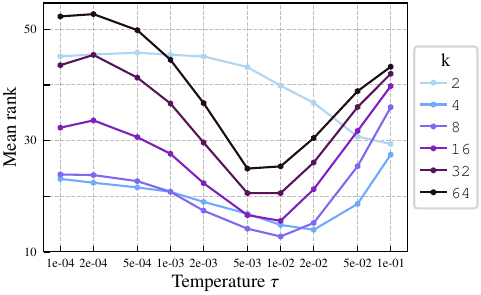}
    \hspace{0.5cm}
    \includegraphics[width=0.49\linewidth]{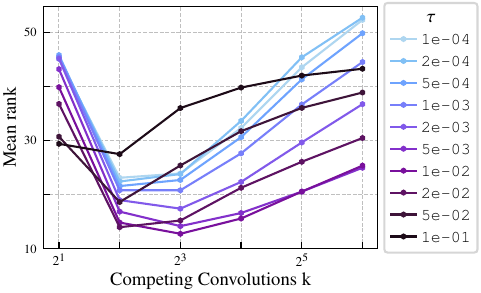}
    \caption{Parameter sensitivity analysis for \texttt{SOCK} on the 112 UCR datasets. The plots display orthogonal slices of the mean rank surface spanned by the softmax temperature $\tau$ and number of competing kernels $k$. \textbf{Top:} Mean rank vs. $\tau$ for fixed $k$. \textbf{Bottom:} Mean rank vs. $k$ for fixed $\tau$. Results are averaged across 30 resamples of the train and test data.}
    \label{fig:k_vs_temp}
\end{figure}

\paragraph{Softmax Temperature $\tau$}
The temperature $\tau$ controls the "softness" of the kernel competition in \texttt{SOCK}. As $\tau \to 0$, the operation resembles something similar to the hard \texttt{argmax} counting used in Hydra; as $\tau \to \infty$, the weights become uniform. We evaluate \texttt{SOCK} for $\tau \in (0.0001, 0.1)$ across different numbers of competing kernels $k \in \{2, 4, 8, 16, 32, 64\}$. We find that if we scale the temperature using the rule $\tau := \log(k) \tau_{base}$, the choice $\tau_{base} = 0.01$ remains optimal on average for all tested values of $k > 2$. These results are illustrated in \cref{fig:k_vs_temp}, using the aforementioned scaling rule.

The case $k=2$ exhibits an interesting phenomenon. Since the softmax weights sum to one, we have that $P_t^{(g,2)} = 1 - P_t^{(g,1)}$. Moreover, because the standard deviation satisfies $\text{std}(X) = \text{std}(1-X)$, the softmax-deviation pooling yields identical features $F^{(g,1)} = F^{(g,2)}$ for each competing kernel pair. Consequently, $k=2$ is a degenerate case where the feature map produces redundant copies rather than capturing competitive dynamics, which explains its comparatively poor performance.

\begin{figure}[h]
    \centering
    \includegraphics[width=0.49\linewidth]{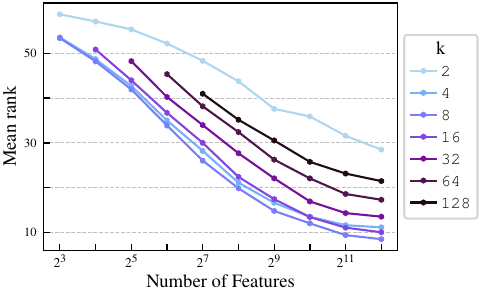}
    \hspace{0.5cm}
    \includegraphics[width=0.49\linewidth]{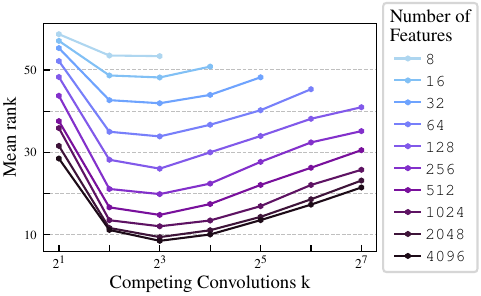}
    \caption{Feature budget analysis on the 112 UCR datasets. The plots correspond to slices of the mean rank surface over the number of competing kernels $k$ and total number of random features (dimension). \textbf{Top:} Mean rank vs. total feature dimension for fixed $k$. \textbf{Bottom:} Mean rank vs. $k$ for fixed total feature budgets. Results are averaged across 30 resamples of the train and test data.}
    \label{fig:k_vs_dim}
\end{figure}

\paragraph{Number of Competing Kernels $k$}
We further ablate the number of competing kernels $k$ within each group of random convolutional kernels. If $g$ is the number of groups, the total number of random features outputted by the feature map is $k \times g$. We investigate which value of $k$ yields the best classification performance given a fixed feature "budget" (target dimension). In \cref{fig:k_vs_dim}, we vary the target dimension and plot the rank of classifiers trained using different values of $k$. Using the scaling rule $\tau := \log(k) \tau_{base}$ established above, we find that $k=8$ yields the best classification accuracy, consistent with the findings for Hydra \citep{2023Hydra}.

\paragraph{Comparison to SOTA}
The left panel of \cref{fig:classification} shows a Critical Difference (CD) diagram comparing \texttt{SOCK} against the top-performing models identified in the recent "Bake Off Redux" benchmark \citep{2024BakeOff}. The diagram presents the average relative rank of each model, indicating statistically significant clusters based on a Wilcoxon signed-rank test with Holm correction. This is a well-established methodology for comparing multiple algorithms across multiple datasets \citep{2006StatisticalComparisonsOfClassifiersOverMultipleDataSets, 2008AnExtensionOnStatisticalComparisonsOfClassifiersOverMultipleDataSetsForAllPairwiseComparisons, 2016ShouldWeReallyUsePostHocTestsBasedOnMeanRanks,2024BakeOff}.

We include the single best model from each category presented in \citep{2024BakeOff}, excluding meta-ensembles (such as HIVE-COTE and HC2) to focus on single-model representations. The selected baselines are: \textbf{Proximity Forest} (Distance), \textbf{FreshPRINCE} (Feature), \textbf{QUANT} (Interval), \textbf{RDST} (Shapelet), \textbf{WEASEL 2.0} (Dictionary), \textbf{MultiRocket} (Random Convolution), and \textbf{H-InceptionTime} (Deep Learning). We also include standard \textbf{Rocket}, \textbf{Hydra}, and \textbf{Signatures} for context. We train a classifier with \texttt{SOCK} features using $k=8$, path augmentations (\texttt{diff}, \texttt{posneg}), and a final number of output features equal to 8192. In the left panel of \cref{fig:classification}, we observe that \texttt{SOCK} achieves the best average rank among all baseline models and is statistically tied with MultiRocket and H-InceptionTime. Crucially, \texttt{SOCK} outperforms Hydra and matches MultiRocket while using an order of magnitude fewer random features and a single pooling operation, rather than four. Moreover, since SOCK features are fully differentiable, this opens up for fine-tuning of random convolutional models via SGD, and inclusion within larger deep neural network architectures. We leave these end-to-end applications for future work as they are outside the scope of this manuscript.

\end{document}